\documentclass{article}

\usepackage[final]{neurips_2025}

\usepackage[utf8]{inputenc} 
\usepackage[T1]{fontenc}    
\usepackage{hyperref}       
\usepackage{url}            
\usepackage{booktabs}       
\usepackage{amsfonts}       
\usepackage{nicefrac}       
\usepackage{microtype}      
\usepackage{xcolor}         
\usepackage{multirow}
\usepackage{array}
\usepackage{graphicx} 
\usepackage{amsmath}  
\usepackage{capt-of}
\usepackage{wrapfig}
\usepackage{caption}
\usepackage{subcaption}  
\usepackage{adjustbox}

\title{Learning Dense Hand Contact Estimation from Imbalanced Data}

\author{
  Daniel Sungho Jung$^{1}$ \hskip1.6em Kyoung Mu Lee$^{1,2}$ \\
   $^{1}$IPAI, $^{2}$Dept. of ECE \& ASRI, Seoul National University, Korea  \\ 
   {\tt\small \{dqj5182, kyoungmu\}@snu.ac.kr} 
}

\begin{document}

\maketitle

\begin{abstract}

Hands are essential to human interaction, and exploring contact between hands and the world can promote comprehensive understanding of their function.
Recently, there have been growing number of hand interaction datasets that cover interaction with object, other hand, scene, and body.
Despite the significance of the task and increasing high-quality data, how to effectively learn dense hand contact estimation remains largely underexplored.
There are two major challenges for learning dense hand contact estimation.
First, there exists class imbalance issue from hand contact datasets where majority of regions are not in contact.
Second, hand contact datasets contain spatial imbalance issue with most of hand contact exhibited in finger tips, resulting in challenges for generalization towards contacts in other hand regions.
To tackle these issues, we present a framework that learns dense HAnd COntact estimation~(HACO) from imbalanced data.
To resolve the class imbalance issue, we introduce balanced contact sampling, which builds and samples from multiple sampling groups that fairly represent diverse contact statistics for both contact and non-contact vertices.
Moreover, to address the spatial imbalance issue, we propose vertex-level class-balanced~(VCB) loss, which incorporates spatially varying contact distribution by separately reweighting loss contribution of each vertex based on its contact frequency across dataset.
As a result, we effectively learn to predict dense hand contact estimation with large-scale hand contact data without suffering from class and spatial imbalance issue.
The codes are available at \url{https://github.com/dqj5182/HACO_RELEASE}.

\end{abstract}

\section{Introduction}
\label{sec:intro}
From infancy, humans rely on physical contact to perceive and interact with the surrounding environment. 
Among the various mediums of contact, hands play a predominant role, facilitating essential tasks and enabling effective communication. 
Consequently, developing a robust hand contact estimation model is crucial for advancing our understanding of hand interactions and addressing various challenges~\cite{grady2021contactopt, liu2024easyhoi} that require accurate hand contact estimation.

Over the recent years, we have witnessed significant advances in large-scale interaction datasets involving hands.
Hand-object interaction datasets~\cite{hasson2019learning, chao2021dexycb, cao2021reconstructing, hampali2020honnotate, hampali2022keypoint, fan2023arctic, liu2022hoi4d, kwon2021h2o} focused on capturing hand grasps of objects.
Hand-hand interaction datasets~\cite{tzionas2016capturing, moon2020interhand2} considered two hand interaction from a single person.
Hand-face interaction dataset~\cite{shimada2023decaf} investigated on interaction between hand and deformable face.
Hand-scene interaction datasets~\cite{hassan2021populating, huang2022capturing, tripathi2023deco} extended the scope of interaction to include environments featuring the ground, walls, and large-scale objects.
Hand-body interaction dataset~\cite{yin2023hi4d} covered broad range of interaction with human body as part of human-human interaction. 
Despite the wide range of diversity in the hand interaction datasets with contact annotations~\cite{hasson2019learning, chao2021dexycb, cao2021reconstructing, hampali2020honnotate, hampali2022keypoint, fan2023arctic, liu2022hoi4d, kwon2021h2o, moon2020interhand2, tzionas2016capturing, shimada2023decaf, hassan2019resolving, huang2022capturing, yin2023hi4d, xie2024rhobin}, to the best of our knowledge, there has been no attempt to build hand contact estimation model trained on such diverse hand contact.
Based on such diverse sets of hand contact data, we aim to build an effective dense hand contact estimation model.

\begin{wrapfigure}{r}{0.5\textwidth}
  \vspace{-2.7mm}
  \begin{center}
  \phantomsubcaption\label{fig:challenges:a}%
  \phantomsubcaption\label{fig:challenges:b}%
  
\includegraphics[width=0.5\textwidth]{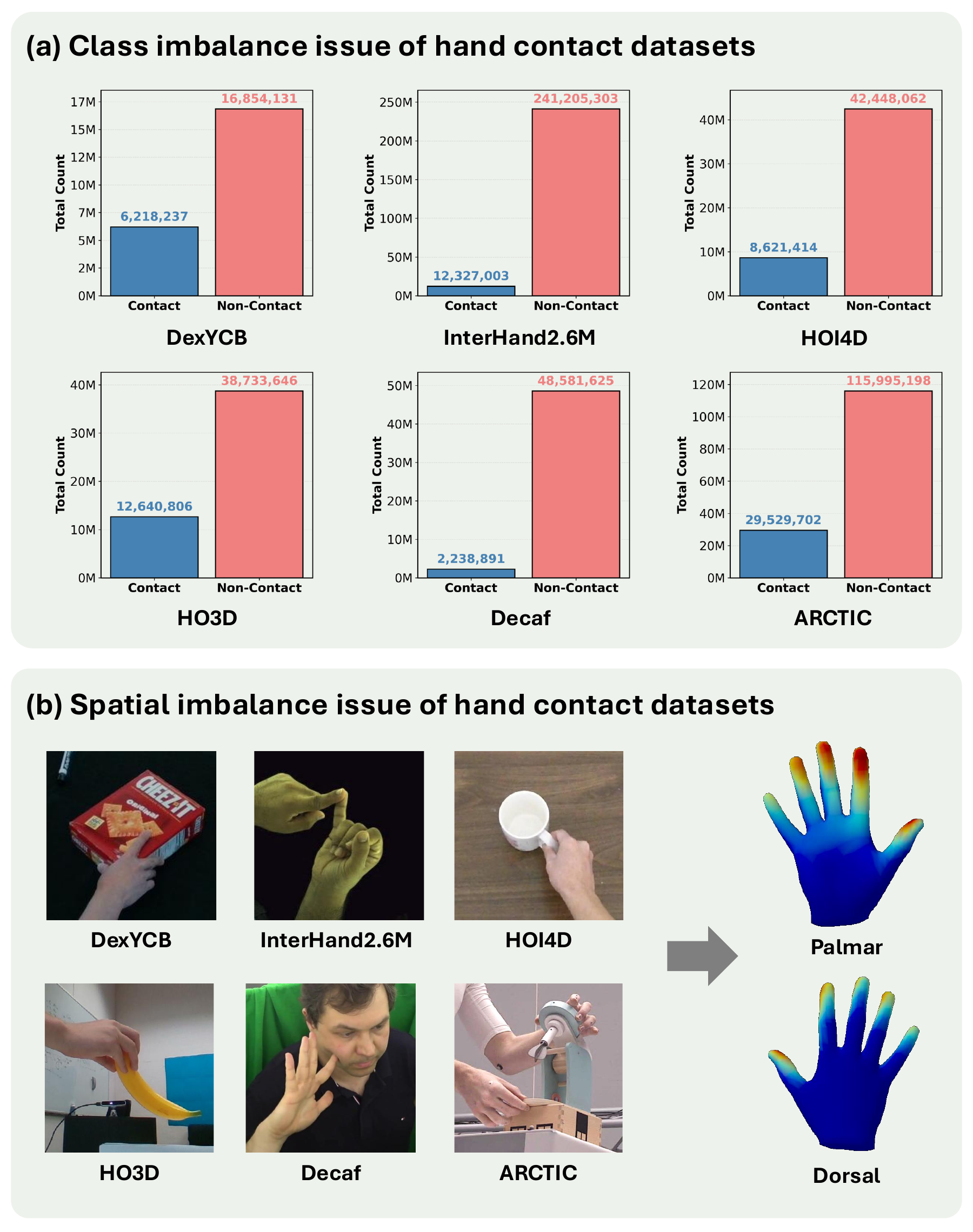}
  \end{center}
  \vspace{-2mm}
  \caption{\footnotesize{\textbf{Two challenges for dense hand contact estimation in the wild.} First, hand contact datasets suffer from class imbalance, as the majority of vertices contain no contact. Second, spatial imbalance arises because contact points are predominantly concentrated at the fingertips. Due to these issues, models trained on such data struggle to generalize to diverse contact patterns across the hand.}}
  \label{fig:challenges}
\end{wrapfigure}

There are two major issues for training a dense hand contact estimation.
First, there exists class imbalance between contact and non-contact class.
Human participants of the hand interaction datasets are often instructed to perform specific actions such as picking up an object~\cite{chao2021dexycb, hampali2020honnotate, liu2022hoi4d} and pointing to specific regions of objects~\cite{kwon2021h2o} or other hands~\cite{moon2020interhand2}.
As most of such actions require precise control with interaction by a small region on hand surface, most of the hand regions are overlooked and excluded from being in contact.
In Figure~\ref{fig:challenges:a}, we provide comparison on the occurrence of contact and non-contact in major hand contact datasets for all vertices of MANO hand model~\cite{romero2017embodied}.
This observation shows severe class imbalance issue of hand contact from existing hand contact datasets.
Numerically, DexYCB~\cite{chao2021dexycb} dataset has roughly 2.7:1 ratio of non-contact to contact vertices.
More severe InterHand2.6M~\cite{moon2020interhand2} dataset has huge imbalance ratio of 19.5:1.
Such imbalance is also shown in Decaf~\cite{shimada2023decaf} dataset with 21.7:1 ratio.
As stated by numerous studies on data imbalance~\cite{cao2019learning, cui2019class, ridnik2021asymmetric, lengpolyloss}, such data imbalance causes poor performance on underrepresented classes, which is in fact the contacting cases in our scenario.
Second, the spatial imbalance issue of hand contact data exists throughout the majority of hand contact datasets. This imbalance arises from the fact that hand interaction predominantly involves the fingertips, which provide a high degree of freedom in both movement and rotation.
Due to such high maneuverability and the precise nature of actions from motion capture datasets, most of hand contact from the datasets are significantly skewed to finger tip regions.
As shown in Figure~\ref{fig:challenges:b}, the heatmap for hand contact from hand contact datasets are highly concentrated at the finger tips.
Due to such spatial imbalance issue of hand contact datasets, models directly trained on the datasets struggle to generalize on diverse contact patterns across hand.

Therefore, we propose HACO, a framework that learns dense hand contact estimation from imbalanced data.
Initially, HACO is trained on a large-scale, assorted dataset composed of 14 datasets, shown in Table~\ref{tab:dataset}, that contains various hand interactions to leverage the power of large-scale training~\cite{brown2020language, rombach2022high, kirillov2023segment, yang2024depth, pavlakos2024reconstructing}.
However, this large-scale dataset still exhibits inherent imbalance in terms of both class and spatial distribution.
To address these issues, we introduce two strategies designed to fully exploit the potential of such large-scale data.
First, we introduce  \textit{Balanced Contact Sampling (BCS)}, which mitigates class imbalance in hand contact datasets by dividing the dataset into multiple sampling groups, each representing different statistics of contact and non-contact.
To ensure fair representation of the contact statistics, we compute a contact balance score that measures how much each hand contact instance deviates from the dataset-wide average.
This metric enables us to assess whether each instance is under-represented or over-represented within the original dataset and to adjust the sampling ratios accordingly, ensuring all contact types are well represented.
Second, we present the  \textit{Vertex-Level Class-Balanced (VCB) loss}, which applies spatially varying weights to each vertex to address the spatial imbalance issue in hand contact datasets.
Inspired by the core idea of class-balanced (CB) loss~\cite{cui2019class}, which re-weights the loss based on class frequency, we adapt this concept to the spatial domain by computing a separate loss weight for each vertex based on how well contact and non-contact are represented at the vertex within the dataset.
While the original CB loss applies global class-level weighting, our method performs fine-grained, vertex-level weighting, allowing the model to focus on under-represented contact patterns in a spatially-aware manner.

As a result, HACO achieves state-of-the-art performance across diverse hand contact scenarios, including hand-object, hand-hand, hand-scene, and hand-body interactions, consistently outperforming existing methods.

Our key contributions are as follows:
\begin{itemize}
\item We introduce HACO, a novel framework that addresses the data imbalance problems in hand contact datasets and enables effective learning of dense hand contact estimation from large-scale data.
\item To mitigate class imbalance in existing hand contact datasets, we propose Balanced Contact Sampling (BCS), which constructs multiple sampling groups to fairly represent diverse contact statistics.
\item To handle spatial imbalance, we present the Vertex-Level Class-Balanced (VCB) loss, which computes a vertex-specific weighting factor based on the dataset-wide contact distribution.
\item In the end, HACO, the first method trained for dense hand contact estimation on a large-scale dataset, achieves substantial performance gains across diverse contact scenarios.
\end{itemize}
\section{Related works}
\label{sec:related_works}

\noindent\textbf{Dense hand contact estimation.}
Most of the existing methods that explore dense hand contact estimation are devoted to dense human-scene contact estimation~\cite{hassan2021populating, huang2022capturing, tripathi2023deco, nam2024joint}.
Given an image of human-scene interaction, such methods estimate contactness for each vertex of 3D human model~(e.g., SMPL~\cite{loper2015smpl}, SMPL-X~\cite{pavlakos2019expressive}).
POSA~\cite{hassan2021populating} is based upon the idea that the human-scene contact is largely influenced by the 3D pose of human.
Hence, the work employs a conditional variational autoencoder framework~\cite{sohn2015learning} to learn human-scene contact with the 3D vertex position of posed human as condition.
BSTRO~\cite{huang2022capturing} takes another approach for estimating human-scene contact, which directly relies on visual input with Transformer-based architecture~\cite{devlin2019bert}.
To train the model, they proposed RICH dataset~\cite{huang2022capturing} that annotates dense human-scene contact by capturing 3D geometry of body and scene with multi-view laser scanner in both indoor and outdoor environment. 
Lastly, to improve generalization ability of human-scene contact estimation, DECO~\cite{tripathi2023deco} focused on how to annotate human-scene contact in the wild.
They introduced simple yet effective contact annotation method that simply asks crowd-sourced annotators to paint 3D human model~(i.e., SMPL~\cite{loper2015smpl}) in neutral T-pose, which significantly enhanced the potential of scaling contact data.
Despite numerous attempts on dense human-scene contact estimation, there lacks in-depth investigation on dense hand contact estimation.
Furthermore, we found that dense hand contact estimation has unique challenges compared to human-scene contact estimation.
First, due to class imbalance of hand contact datasets, the distribution of the datasets are often skewed towards non-contact class, making it non-trivial to train hand contact estimation from the datasets.
Second, there exists spatial imbalance issue of hand contact datasets where majority of hand contact occurs in finger tips with insufficient data for hand contact in rest of the regions~(e.g., dorsum).
We introduce HACO, a dense hand contact estimation method that effectively tackles such data imbalance issues of hand contact datasets.

\begin{table*}[t]
\centering
\small
\caption{\textbf{Dataset configuration.} We leverage 14 datasets with various hand interaction.}
\scalebox{0.87}{\begin{tabular}{lccccc} \toprule
Dataset & Interaction & Domain & \# of images & \# of subjects & \# of objects \\
\midrule
ObMan~\cite{hasson2019learning} & Rigid Object & Synthetic & 152K & 20 & 8 \\
DexYCB~\cite{chao2021dexycb} & Rigid Object & Real / Indoor & 582K & 10 & 20\\
MOW~\cite{cao2021reconstructing} & Rigid Object & Real / Outdoor & 0.5K & 450 & 121 \\
HO3D~\cite{hampali2020honnotate} & Rigid Object & Real / Indoor & 78K & 10 & 10 \\
H2O3D~\cite{hampali2022keypoint} & Rigid Object + Hand & Real / Indoor & 76K & 5 & 10 \\
ARCTIC~\cite{fan2023arctic} & Articulated Object + Hand & Real / Indoor / Studio & 2.1M & 10 & 11 \\
HOI4D~\cite{liu2022hoi4d} & Rigid Object + Hand & Real / Indoor & 2.4M & 4 & 16 \\
H2O~\cite{kwon2021h2o} & Rigid Object + Hand & Real / Indoor & 571K & 4 & 8 \\
InterHand2.6M~\cite{moon2020interhand2} & Hand & Real / Indoor / Studio & 2.6M & 27 & -- \\
HIC~\cite{tzionas2016capturing} & Hand & Real / Indoor & 36K  & 1 & -- \\
PROX~\cite{hassan2019resolving} & Scene & Real / Indoor / Studio & 100K & 20 & -- \\
RICH~\cite{huang2022capturing} & Scene & Real / Outdoor & 540K & 22 & -- \\
Decaf~\cite{shimada2023decaf} & Face (Whole-Body) &  Real / Indoor & 100K & 8 & -- \\
Hi4D~\cite{yin2023hi4d} & Whole-Body & Real / Indoor / Studio & 11K & 40 & -- \\

\bottomrule
\end{tabular}}
\vspace{-0.3cm}
\label{tab:dataset}
\end{table*}

\noindent\textbf{3D hand and object interaction reasoning.}
Dense hand contact reveals key spatial and semantic relationship between 3D hand and object.
Initially, dense hand contact was studied in the field of grasping with specific tasks such as grasp contact prediction~\cite{brahmbhatt2019contactdb}, contact-guided hand and object pose refinement~\cite{grady2021contactopt}, and contact-conditioned grasp generation~\cite{liu2023contactgen}.
Such works made tremendous impact on robotics, especially on how robots can naturally grasp an object.
However, their dense hand contact were mostly predicted from 3D geometry of hand and object without exploring how to directly predict the dense hand contact from visual input.
Recently, the field of 3D hand and object reconstruction showed potential on the effectiveness of contact for improving both reconstruction accuracy and natural interaction between hand and object.
ContactOpt~\cite{grady2021contactopt} proposed a differentiable contact optimization that optimizes hand and object poses inferred from image-based methods with estimated contact to achieve natural interaction.
EasyHOI~\cite{liu2024easyhoi} introduced a prior-guided hand-object interaction optimizer that refines initial 3D hand and object using heuristically determined contact regions to improve interaction quality.
Our HACO is applicable to these downstream tasks as long as visual input is available.
To demonstrate HACO’s effectiveness, we show that our predicted dense contact improves 3D grasp optimization over geometry-driven contact estimation from ContactOpt~\cite{grady2021contactopt}, and 3D reconstruction over the previously used heuristic contact regions in EasyHOI~\cite{liu2024easyhoi}.

\section{Method}

\begin{figure*}[t]
\begin{center}
\includegraphics[width=1.0\linewidth]{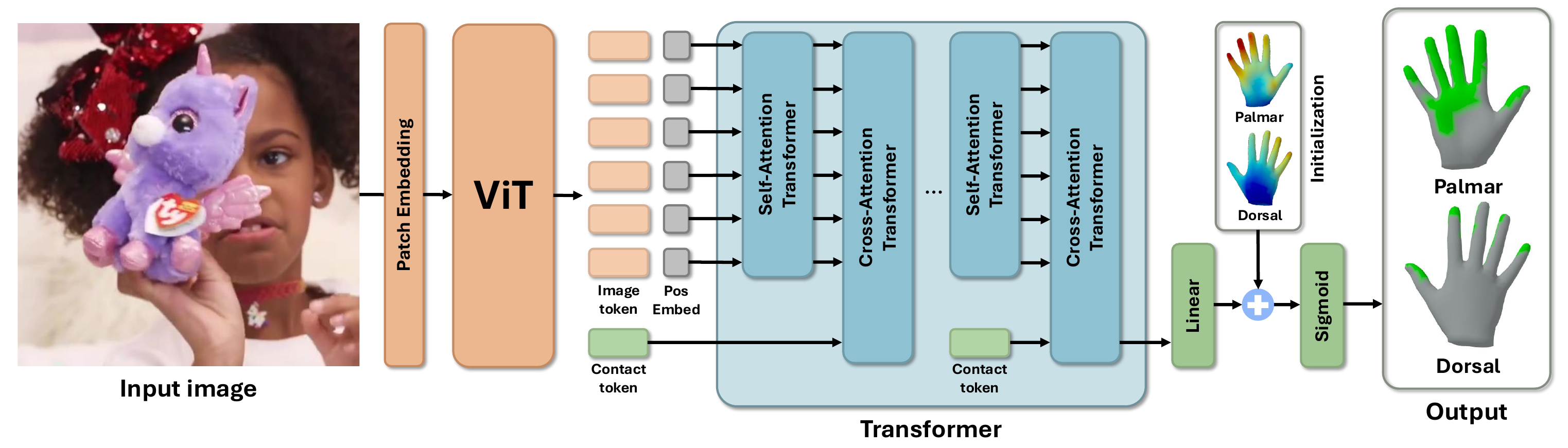}
\end{center}
\vspace{-3.5mm}
\caption{
\textbf{Overall pipeline of HACO.} Our method encodes input image as image tokens with a ViT backbone after patch embedding layers. Given the image tokens along with positional embeddings and a contact token, multiple layers of self-attention Transformer and cross-attention Transformer produce an output token.
Lastly, the output token is passed through a linear layer and combined with the contact initialization, followed by a sigmoid activation to output the final hand contact prediction.}
\label{fig:pipeline}
\vspace{-0.3cm}
\end{figure*}

\subsection{Model architecture}
Given an RGB image~$\mathbf{I} \in \mathbb{R}^{3 \times H \times W}$, where $H$ and $W$ denote the height and width of the image, we first embed the image into multiple tokens with a patch embedding layer and process these tokens using a Vision Transformer~(ViT)~\cite{dosovitskiy2020image}.
Following ViTPose~\cite{xu2022vitpose} and HaMeR~\cite{pavlakos2024reconstructing}, we reshape the tokens and formulate the image feature~$\mathbf{F} \in \mathbb{R}^{1280 \times 16 \times 12}$.
Additionally, we build a contact token, which acts as a query token for hand contact estimation.
To inject image feature~$\mathbf{F}$ into the contact token, we process the image feature~$\mathbf{F}$ and positional embedding with consecutive self-attention Transformer and cross-attention Transformer with contact token as query and image feature~$\mathbf{F}$ as key-value pair.
Afterwards, the output contact token is processed with linear layer.
In order to resemble residual layers~\cite{he2016deep} that stabilize training, we add contact initialization to the output of linear layer.
The contact initialization is a learnable embedding that naturally learns the most effective initial contact estimation during training.
Lastly, a sigmoid layer converts the final contact logits~$\mathbf{C} \in \mathbb{R}^{V}$ into the estimated dense hand contact probabilities, where $V = 778$ is MANO hand vertices~\cite{romero2017embodied}.

\subsection{Balanced contact sampling}
\label{sec:main_bcs}
A hand contact dataset~$\mathbf{D} = \{ \mathbf{H}_i = (\mathbf{v}_i, \mathbf{c}_i) \}_{i=1}^N$ consists of $N$ hand instances, where each $\mathbf{H}_i$ contains the 3D hand vertices $\mathbf{v}_i \in \mathbb{R}^{V \times 3}$ and a binary vertex contact vector $\mathbf{c}_i \in \{0,1\}^V$.
As illustrated in Figure~\ref{fig:challenges:a}, the dataset~$\mathbf{D}$ exhibits a strong class imbalance, with many hand instances lacking contact.  
To alleviate this issue, we propose balanced contact sampling (BCS), which constructs sampling bins~$\mathbf{B} = \{ \mathcal{B}_1, \mathcal{B}_2, \dots, \mathcal{B}_K \}$ consisting of $K$ groups of hand instances.  
Each bin~$\mathcal{B}_k$ is formed by binning samples based on their contact balance scores~$s_i$, which reflect how each hand's contact~$\mathbf{c}_i$ deviates from the dataset-wide average~$\bar{\mathbf{c}}$.  
To compute $s_i$, we first define the dataset-wide contact mean as $\bar{\mathbf{c}} = \frac{1}{N} \sum_{i=1}^N \mathbf{c}_i$, which represents the contact probability at each vertex across all samples.  
The contact balance score~$s_i$ for each sample is then defined as:
\begin{equation}
\label{eq:cb_score}
s_i = \frac{1}{V} \left( \mathbf{c}_i^\top (1 - \bar{\mathbf{c}}) - \mathbf{c}_i^\top \bar{\mathbf{c}} \right)
\end{equation}
where higher scores correspond to hand instances~$\mathbf{H}_i$ whose contact patterns deviate more from the dataset-wide average, enabling balanced grouping across diverse contact statistics that span both contact and non-contact samples.
Based on the distribution of contact balance scores $\{s_i\}_{i=1}^N$, we partition the dataset into $K$ bins~$\mathcal{B}_1, \dots, \mathcal{B}_K$ by applying non-linear binning.  
We first compute the minimum and maximum contact balance scores as $s_{\min} = \min_i s_i$ and $s_{\max} = \max_i s_i$, and define $K{+}1$ bin edges using a logarithmic spacing scheme controlled by a curvature parameter $\beta > 0$ (e.g., $\beta = 5$):
\begin{equation}
\tau_k = s_{\min} + (s_{\max} - s_{\min}) \cdot \frac{\log(1 + \beta \cdot x_k)}{\log(1 + \beta)}, \quad
x_k = \frac{k}{K}, \quad k = 0, \dots, K.
\end{equation}
This results in bin edges $\{ \tau_0, \tau_1, \dots, \tau_K \}$ that allocate finer resolution to higher contact scores.  
Each group~$\mathcal{B}_k$ is then constructed by selecting hand instances whose contact balance scores fall within the corresponding interval:
\begin{equation}
\mathcal{B}_k = \left\{ \mathbf{H}_i \in \mathbf{D} \,\middle|\, s_i \in [\tau_{k-1}, \tau_k) \right\}
\end{equation}
Due to the skewed distribution of scores~${s_i}$, the bins~${\mathcal{B}_k}$ may differ in size. To equalize their contribution, we apply stratified resampling to ensure each bin contains the same number of hand instances in the final training set.
In the end, this binning strategy enables more accurate grouping and fair sampling of contact-rich hand instances that are underrepresented in the original dataset~$\mathbf{D}$, improving the representational balance of the sampling bins~$\mathbf{B}$.

\subsection{Vertex-level class-balanced loss}
\label{sec:main_vcb_loss}
As illustrated in Figure~\ref{fig:challenges:b}, hand contact datasets exhibit severe spatial imbalance, with contact heavily concentrated at the fingertips.
To address this issue, we propose a vertex-level class-balanced (VCB) loss, which builds upon the class-balanced (CB) sigmoid binary cross-entropy loss~\cite{cui2019class}.
We first formulate hand contact estimation as a binary classification problem.
A simple binary cross-entropy (BCE) loss is defined as $\ell_{\text{BCE}}(y, p) = -y \log(p) - (1 - y) \log(1 - p)$, where $y \in \{0, 1\}$ is the ground-truth label and $p \in [0, 1]$ is the predicted probability.
When applied over all hand vertices~$V$, the overall BCE loss becomes:
\begin{equation}
\label{eq:bce_loss}
\mathcal{L}_{\text{BCE}} = \frac{1}{|V|} \sum_{v \in V} \ell_{\text{BCE}}(y_v, p_v),
\end{equation}
where we denote ground-truth contact for vertex~$v$ as $y_v \in \{ 0, 1 \}$ and predicted contact probability for vertex~$v$ as $p_v = \sigma(z)$ with sigmoid function as $\sigma$ and predicted logit as $z$.

From the BCE loss, class-balanced (CB) loss~\cite{cui2019class} introduces a weighting factor~$\alpha_c$ for each class~$c$, assigning greater importance to under-represented classes.
In our case, the class~$c$ corresponds to the binary contact label~$y \in \{0, 1 \}$.
Accordingly, we denote the class-balanced weighting factor as~$\alpha_y$.
Simply, we can write CB loss as follows:
\begin{equation}
\mathcal{L}_{\text{CB}} = \frac{1}{|V|} \sum_{v \in V} \alpha_{y} \ell_{\text{BCE}}(y_v, p_v).
\end{equation}
Specifically, the weighting factor~$\alpha_y$ is defined using the effective number of samples~$E_n^{(y)}$, where
\begin{equation}
\alpha_y = \frac{1}{E_n^{(y)}} = \frac{1 - \beta}{1 - \beta^{n_y}}.
\end{equation}
Here, $n_y$ denotes the number of occurrences of class~$y$, and $\beta \in [0, 1)$ is a hyperparameter that controls the influence of sample count on the weighting.  
When the contact class~$y$ appears frequently in the dataset, the contact frequency value $n_y$ becomes large, which increases the effective number~$E_n^{(y)} = \frac{1 - \beta^{n_y}}{1 - \beta}$ and results in a smaller class-balanced weight~$\alpha_y = \frac{1}{E_n^{(y)}}$.  
As a result, the CB loss down-weights the contribution of frequent classes and helps mitigate class imbalance in dense hand contact datasets.

Nevertheless, CB loss~\cite{cui2019class} is defined at the class level, based on the binary contact label~$\mathbf{y} \in \{ 0, 1 \}$.  
If we directly apply CB loss, the weighting factor~$\alpha_y$ takes only two distinct values, one for contact and one for non-contact, and is applied uniformly across all spatial regions.  
In the DexYCB dataset~\cite{chao2021dexycb}, contact and non-contact vertices account for approximately 76.86\% and 23.14\% of all vertex labels, respectively.  
Accordingly, the weighting factor~$\alpha_y$ is computed using sample counts~$n_y = (0.7686 \times n, 0.2314 \times n)$, where $n$ is the total number of contact labels in the dataset.  
However, this count~$n_y$ is applied uniformly across all vertices, without considering spatial variation.  
For example, fingertip vertices are much more likely to be in contact, while dorsal hand vertices are rarely in contact.  
As a result, the class-balanced weight~$\alpha_y$ and effective sample number~$E_n^{(y)}$ do not accurately reflect the local distribution of contact and non-contact at each vertex.  
To address this limitation, we extend the CB loss formulation to operate at the vertex level.

We introduce the vertex-level class-balanced (VCB) loss, which computes a spatially varying weighting factor~$\alpha_{y_v}$ and effective number of samples~$E_n^{(y_v)}$ for each vertex~$v$.  
To account for the spatial imbalance in hand contact datasets, we simply assign a separate weighting factor~$\alpha_{y_v}$ to each vertex, depending on the local distribution of contact labels.  
Given the number of samples~$n_{y_v}$ for contact class~$y$ at vertex~$v$, the weighting factor is computed as:
\begin{equation}
\alpha_{y_v} = \frac{1}{E_n^{(y_v)}} = \frac{1 - \beta}{1 - \beta^{n_{y_v}}}.
\end{equation}
Using this spatially varying weight, the vertex-level class-balanced loss is defined as:
\begin{equation}
\mathcal{L}_{\text{VCB}} = \frac{1}{|V|} \sum_{v \in V} \alpha_{y_v} \, \ell_{\text{BCE}}(y_v, p_v).
\end{equation}
This spatial adaptivity enables the VCB loss to effectively mitigate the spatial imbalance inherent in hand contact datasets.
The VCB loss is applied through a progressive weighting strategy that combines the CB loss and VCB loss throughout training. 
At the beginning of training, only the CB loss is applied, 
while the weight of the VCB component increases linearly with each epoch. 
The VCB weight reaches a fixed maximum at the final epoch, resulting in a smooth transition from global to vertex-adaptive supervision.
Lastly, we include a lightweight false-positive regularization term to suppress overly confident contact predictions for non-contact vertices.

\subsection{Final outputs and loss functions}
The final dense hand contact output supports multi-level supervision through projections to coarser resolutions.
From the dense contact logits~$\mathbf{C}$ regressed by the final linear layer, we build multi-level hand contact predictions~$\mathbf{C'} = \{ \mathbf{C}_i = \mathcal{J}_i^{v_i}\times \mathbf{C} \}_{i=1}^N$ for~$\mathbf{N} = 4$ and~$\mathbf{v_i} \in \{778, 336, 84, 21\}$ where $\mathcal{J}_i^{v_i} \in \mathbb{R}^{{v_i} \times V}$ is a regressor that maps full MANO vertices into coarse representations.
We then apply a sigmoid layer to convert the logits into contact probabilities. 
To train HACO, we apply three loss functions: the vertex-level class-balanced (VCB) loss for multi-level hand contact supervision, a regularization loss, and a smoothness loss. 
Please refer to Section~\ref{sec:main_vcb_loss} for details on the VCB loss, and to the Appendix for the regularization and smoothness losses. 
Briefly, the regularization loss is defined as the L1 loss between the dense contact prediction and the dataset-wide mean ground-truth contact, encouraging the prediction to stay close to the average. 
The smoothness loss measures how predicted contact and non-contact regions are spatially disconnected. 
This encourages HACO to predict a small number of large contact regions rather than many small, fragmented ones. 
We set the weights of the VCB loss, regularization loss, and smoothness loss to 1, 0.1, and 1, respectively.

\section{Implementation details}
PyTorch~\cite{paszkepytorch} is used for implementation.
Our backbone is initialized with the pre-trained weights of publicly released HaMeR~\cite{pavlakos2024reconstructing}.
Following HaMeR, we apply data augmentations including random scaling, cropping, and rotation. 
To improve robustness to degraded inputs, we additionally perform low-resolution, noise, and blur augmentations. 
We use the AdamW optimizer~\cite{loshchilov2018decoupled} with a learning rate of $10^{-5}$ and a mini-batch size of 24. 
For stable convergence, the learning rate is reduced by a factor of 0.9 after 5 and 10 epochs. 
We train HACO for 10 epochs on a single NVIDIA A6000 GPU.
\section{Experiments}

\subsection{Datasets}
We select 14 datasets with diverse hand interactions including ObMan~\cite{hasson2019learning}, DexYCB~\cite{chao2021dexycb}, MOW~\cite{cao2021reconstructing}, HO3D~\cite{hampali2020honnotate}, H2O3D~\cite{hampali2022keypoint}, ARCTIC~\cite{fan2023arctic}, HOI4D~\cite{liu2022hoi4d}, H2O~\cite{kwon2021h2o} for hand-object interaction, InterHand2.6M~\cite{moon2020interhand2}, HIC~\cite{tzionas2016capturing} for hand-hand interaction, PROX~\cite{hassan2019resolving}, RICH~\cite{huang2022capturing} for hand-scene interaction, and Decaf~\cite{shimada2023decaf}, Hi4D~\cite{yin2023hi4d} for hand-body interaction.
To reduce redundancy from large video datasets, we employ sampling ratio of 5, 10, 5 for HOI4D~\cite{liu2022hoi4d}, InterHand2.6M~\cite{moon2020interhand2}, and Decaf~\cite{shimada2023decaf} dataset, respectively.
Note that this is separate from the balanced contact sampling in Section~\ref{sec:main_bcs}.
For ARCTIC~\cite{fan2023arctic} dataset, we select data captured from an egocentric viewpoint, while for HOI4D~\cite{liu2022hoi4d}, we use data involving rigid objects due to their uniqueness.
Since the 3D hand annotations in PROX~\cite{hassan2019resolving} are inaccurate in the qualitative split, we use only the quantitative subset. For the RICH dataset, we follow the official split used in BSTRO~\cite{huang2022capturing} for fair comparison.
In total, this results in 655K images with ground-truth dense hand contact labels.
For details on our method on extracting ground-truth dense hand contact labels, please refer to Appendix.

\subsection{Evaluation metrics}
To evaluate dense hand contact estimation, we compute precision, recall, and F1-score.
However, recall and F1-score are undefined for fully non-contact samples.
Thus, we skip fully non-contacting hands during evaluation.
We also assess contact quality on two downstream tasks: 3D hand grasp optimization~\cite{grady2021contactopt} and 3D hand-object reconstruction~\cite{liu2024easyhoi}.
For 3D hand grasp optimization~\cite{grady2021contactopt}, we evaluate Intersection Volume~(Inter Vol.), Mean Per-Joint Position Error~(MPJPE), and contact metrics.
For 3D hand-object reconstruction, we compare HACO using Mean Per-Vertex Position Error~(MPVPE), MPJPE, $\text{CD}_{\text{ho}}$, $\text{F-5}_{\text{ho}}$, and $\text{F-10}_{\text{ho}}$ against the contact method of EasyHOI.
$\text{F-5}_{\text{ho}}$ and $\text{F-10}_{\text{ho}}$ denote F-scores of reconstructed 3D hand and object using 5mm and 10mm thresholds.

\subsection{Ablation study}

\noindent\textbf{Effectiveness of balanced contact sampling.}
In Table~\ref{tab:abl_data_imb_sampling}, the proposed balanced contact sampling strategy significantly enhances the performance across all metrics. Compared to the model trained without sampling, our method improves 1.0\% in precision, 12.0\% in recall, and 8.5\% in F1-score.
\begin{wraptable}{r}{0.5\linewidth}
\vspace{-0.2cm}
\centering
\small
\setlength{\tabcolsep}{2pt}  
\caption{\textbf{Ablation on balanced contact sampling strategy on MOW~\cite{cao2021reconstructing} dataset.}}
\begin{tabular}{lcccc} \toprule
Methods & $\text{Precision}{\uparrow}$ & $\text{Recall}{\uparrow}$ & $\text{F1-Score}{\uparrow}$ \\ 
\midrule
w/o sampling & 0.520 & 0.542 & 0.481 \\
w/ sampling~(Ours) & \textbf{0.525} & \textbf{0.607} & \textbf{0.522} \\ 
\bottomrule
\end{tabular}
\label{tab:abl_data_imb_sampling}
\vspace{-0.3cm}
\end{wraptable}
The substantial boost in recall, in particular, highlights the effectiveness of balanced sampling in mitigating class imbalance by exposing the model to more positive contact instances. 
Overall, these results demonstrate that our sampling strategy plays a critical role in improving our model’s ability to detect contact accurately and comprehensively.

\noindent\textbf{Effectiveness of vertex-level class-balanced loss.}
In Table~\ref{tab:abl_data_imb_loss}, the proposed VCB loss achieves the best overall performance by effectively addressing data imbalance at a finer granularity.
\begin{wraptable}{r}{0.55\linewidth}
\vspace{-0.2cm}
\centering
\small
\setlength{\tabcolsep}{4pt}  
\caption{\textbf{Comparison of various loss-based techniques for data imbalance on MOW~\cite{cao2021reconstructing} dataset.}}
\begin{tabular}{lcccc} \toprule
Methods & $\text{Precision}{\uparrow}$ & $\text{Recall}{\uparrow}$ & $\text{F1-Score}{\uparrow}$ \\ 
\midrule
CE loss & 0.530 & 0.294 & 0.348 \\
L1 loss & 0.521 & 0.392 & 0.413 \\
L2 loss & \underline{0.531} & 0.298 & 0.352 \\
Focal loss~\cite{lin2017focal} & 0.518 & 0.387 & 0.409\\ 
CB loss~\cite{cui2019class} & 0.484 & \underline{0.534} & \underline{0.465} \\
CB Focal loss~\cite{cui2019class} & 0.522 & 0.360 & 0.392 \\
LDAM loss~\cite{cao2019learning} & \textbf{0.532} & 0.224 & 0.293 \\
Asymmetric loss~\cite{ridnik2021asymmetric} & 0.484 & 0.479 & 0.440 \\
Poly loss~\cite{lengpolyloss} & 0.528 & 0.324 & 0.371 \\
VCB loss~(Ours) & 0.525 & \textbf{0.607} & \textbf{0.522} \\
\bottomrule
\vspace{-0.6cm}
\end{tabular}
\label{tab:abl_data_imb_loss}
\end{wraptable}
Unlike CB loss, which applies class balancing at the hand-level by weighting positive and negative samples across all vertices, VCB loss operates at the per-vertex level, allowing the model to learn contact patterns from individual vertex more accurately. 
This vertex-wise weighting enables the model to better capture subtle and spatially varying contact signals, especially in under-represented regions, leading to improved recall without sacrificing precision. 
As a result, VCB loss yields the highest F1-score among all loss-based methods, demonstrating that resolving class imbalance locally, rather than globally, is key to effective dense contact estimation.

\begin{table}[t]
\centering
\small
\setlength{\tabcolsep}{4pt}  
\caption{\textbf{Comparison of various training dataset configurations on MOW~\cite{cao2021reconstructing}, HIC~\cite{tzionas2016capturing}, RICH~\cite{huang2022capturing}, Hi4D~\cite{yin2023hi4d}.} HO: Hand-object, HH: Hand-hand, HS: Hand-scene, HB: Hand-body.}
\begin{tabular}{llccc}
\toprule
Test dataset & Train dataset & Precision $\uparrow$ & Recall $\uparrow$ & F1-Score $\uparrow$ \\
\midrule
\multirow{4}{*}{MOW~\cite{cao2021reconstructing}} 
& HS & 0.397 & 0.516 & \underline{0.411} \\
& HS+HH & 0.405 & 0.420 & 0.373 \\
& HS+HH+HB & \underline{0.425} & \underline{0.551} & 0.408 \\
& HS+HH+HB+HO (Ours) & \textbf{0.525} & \textbf{0.607} & \textbf{0.522} \\
\midrule
\multirow{4}{*}{HIC~\cite{tzionas2016capturing}} 
& HS & \underline{0.188} & 0.231 & \underline{0.187} \\
& HS+HO & 0.094 & 0.260 & 0.134 \\
& HS+HO+HB & 0.114 & \underline{0.265} & 0.158 \\
& HS+HO+HB+HH (Ours) & \textbf{0.216} & \textbf{0.409} & \textbf{0.263} \\
\midrule
\multirow{4}{*}{RICH~\cite{huang2022capturing}} 
& HO & 0.551 & 0.547 & \underline{0.492} \\
& HO+HH & \underline{0.554} & 0.522 & 0.471 \\
& HO+HH+HB & \underline{0.554} & \underline{0.581} & 0.490 \\
& HO+HH+HB+HS (Ours) & \textbf{0.741} & \textbf{0.899} & \textbf{0.781} \\
\midrule
\multirow{4}{*}{Hi4D~\cite{yin2023hi4d}} 
 & HS & 0.515 & 0.439 & 0.419 \\
 & HS+HO & \textbf{0.610} & 0.503 & 0.496 \\
 & HS+HO+HH & \underline{0.596} & \underline{0.513} & \underline{0.498} \\
 & HS+HO+HH+HB (Ours) & 0.555 & \textbf{0.636} & \textbf{0.565} \\
\bottomrule
\end{tabular}
\label{tab:abl_data_all}
\vspace{-0.4cm}
\end{table}

\noindent\textbf{Effectiveness of large-scale data configuration.}
Our final model benefits significantly from the inclusion of all four interaction types: hand-object (HO), hand-hand (HH), hand-body (HB), and hand-scene (HS).
As shown in Table~\ref{tab:abl_data_all}, performance improves consistently across all evaluated datasets as more interaction types are incrementally incorporated, with the full combination achieving the highest overall F1-score.
Removing HO leads to a substantial drop in recall, further highlighting its essential role in capturing diverse contact patterns.
Although HB and HH alone contribute modestly, their addition enhances F1-scores and stabilizes overall performance, indicating their complementary effects.
Omitting HS results in reduced F1-score despite strong contributions from the other components, suggesting its importance in separating hand contacts from background context.
Ultimately, only the full configuration achieves strong and balanced performance across all metrics and datasets, demonstrating that each interaction type contributes uniquely and that leveraging the full diversity is crucial for learning accurate and generalizable contact representations.

\subsection{Comparison with state-of-the-art methods}

\begin{figure*}[t]
\begin{center}
\includegraphics[width=1.0\linewidth]{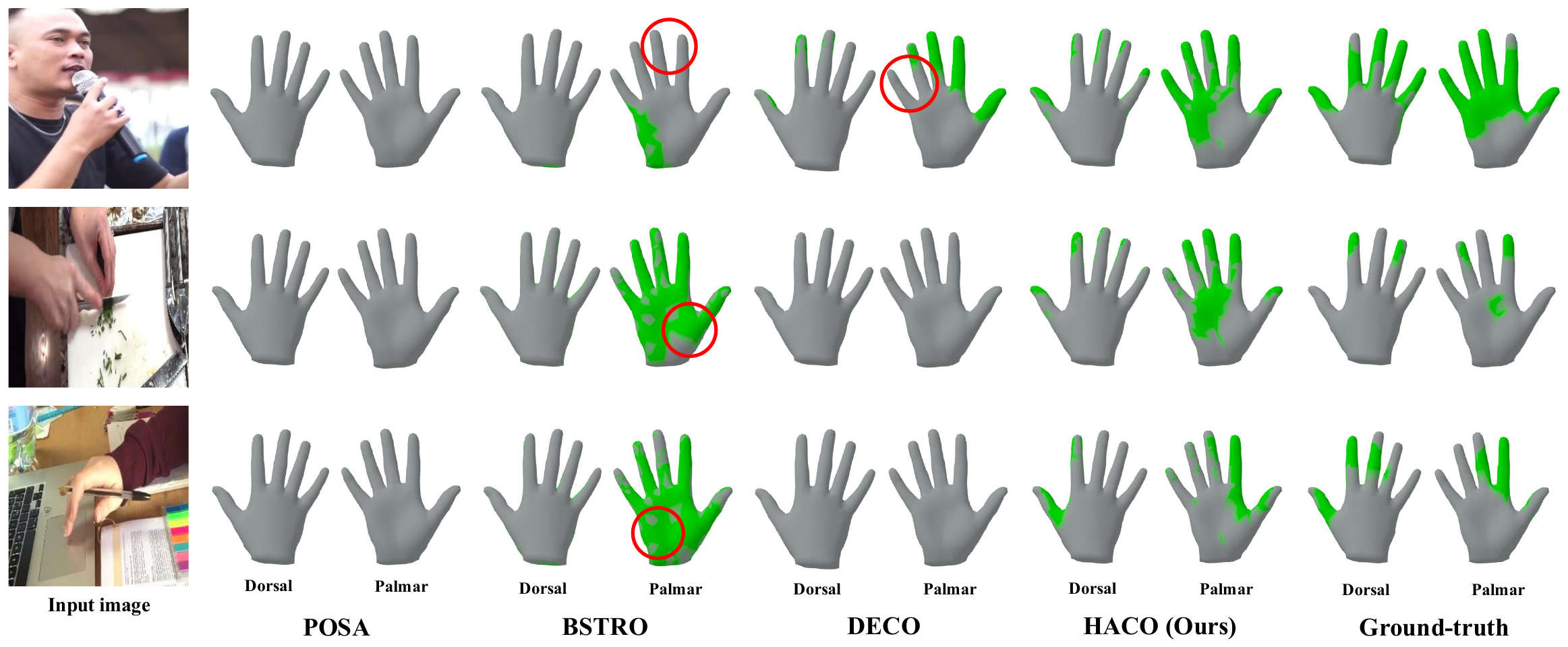}
\end{center}
\vspace{-3.5mm}
\caption{\textbf{Qualitative comparison of dense hand contact estimation with POSA~\cite{hassan2021populating}, BSTRO~\cite{huang2022capturing}, DECO~\cite{tripathi2023deco} on MOW~\cite{cao2021reconstructing} dataset.} We highlight exemplar regions where HACO outperforms previous methods. Note that we only predict right hand contact.}
\label{fig:sota_qual_hand_contact}
\vspace{-0.3cm}
\end{figure*}

\noindent\textbf{Dense hand contact estimation.}
Table~\ref{tab:sota_hand_cont} presents a comparison between our HACO and state-of-the-art approaches of POSA~\cite{hassan2021populating}, BSTRO~\cite{huang2022capturing}, and DECO~\cite{tripathi2023deco} on the MOW~\cite{cao2021reconstructing} dataset.
\begin{wraptable}{r}{0.45\linewidth}
\vspace{-0.2cm}
\centering
\small
\setlength{\tabcolsep}{2pt}  
\caption{\textbf{Comparison with SOTA methods of hand contact estimation on MOW~\cite{cao2021reconstructing} dataset.}}
\begin{tabular}{lcccc} \toprule
Methods & $\text{Precision}{\uparrow}$ & $\text{Recall}{\uparrow}$ & $\text{F1-Score}{\uparrow}$ \\ 
\midrule 
POSA~\cite{hassan2021populating} & 0.134 & 0.128 & 0.101\\ 
BSTRO~\cite{huang2022capturing} & 0.204	& 0.126 & 0.112 \\ 
DECO~\cite{tripathi2023deco} & \underline{0.246} & \underline{0.235} & \underline{0.197} \\ 
\midrule
HACO~(Ours) & \textbf{0.525} & \textbf{0.607} & \textbf{0.522} \\ 
\bottomrule
\vspace{-0.6cm}
\end{tabular}
\label{tab:sota_hand_cont}
\end{wraptable}
Our method achieves substantial improvements across all metrics, significantly outperforming the strongest prior method, DECO~\cite{tripathi2023deco}.
POSA~\cite{hassan2021populating} shows limited performance due to its sole reliance on pose priors~\cite{feng2021collaborative} without image evidence, while BSTRO~\cite{huang2022capturing} suffers from the lack of large-scale, diverse contact data and does not address class or spatial imbalance during training.
DECO~\cite{tripathi2023deco} improved over earlier methods, but performance remains constrained by the lack of diverse supervision and mechanisms for handling imbalance.
In contrast, HACO addresses these challenges by training on diverse contact configurations, initializing contact prediction with a learned prior, and tackling class and spatial imbalance through targeted loss design.
Figure~\ref{fig:sota_qual_hand_contact} further highlights HACO’s qualitative superiority, showing more precise and anatomically plausible contact patterns across scenarios, such as fingertip and palmar contact when holding a mic, partial palmar contact when dicing with knife, and thumb–index finger contact during grasp of a pencil.
Overall, our method shows superior performance over prior SOTA methods, demonstrating its potential as a strong model for dense hand contact estimation by addressing key challenges that limit earlier approaches.

\noindent\textbf{3D hand grasp optimization.}
Table~\ref{tab:sota_grasp_opt} compares our HACO with the prior state-of-the-art method, DeepContact~\cite{grady2021contactopt}, for contact-guided 3D grasp optimization using predicted hand-object poses from HFL-Net~\cite{lin2023harmonious} on the DexYCB~\cite{chao2021dexycb} dataset.
Despite DeepContact having access to full 3D hand and object meshes for contact estimation, our method achieves comparable or superior results using only image input.
HACO attains lower MPJPE, which indicates more accurate hand articulation. 
Furthermore, it achieves higher precision, recall, and F1-score, demonstrating improved contact quality and coverage.
Overall, these results show that HACO enables more effective and robust contact-aware grasp optimization, even without access to full 3D hand and object geometry, highlighting its practical applicability in real-world scenarios.

\begin{table}[htbp]
\vspace{-0.2cm}
\centering
\scriptsize
\setlength{\tabcolsep}{2pt}  
\caption{\textbf{Comparison of various contact estimation methods for 3D hand grasp optimization on DexYCB~\cite{chao2021dexycb} dataset.}}
\begin{tabular}{lcccccc} \toprule
Methods & $\text{Inter Vol.}{\downarrow}$ & $\text{MPJPE}{\downarrow}$ & $\text{Precision}{\uparrow}$ & $\text{Recall}{\uparrow}$ & $\text{F1-Score}{\uparrow}$ \\ 
\midrule
DeepContact~\cite{grady2021contactopt} & \textbf{29.175} & 37.155 & 0.522 & 0.830 & 0.612 \\ 
\midrule
HACO~(Ours) & 29.264 & \textbf{36.520} & \textbf{0.558} & \textbf{0.877} & \textbf{0.666} \\ 
\bottomrule
\vspace{-0.5cm}
\end{tabular}
\label{tab:sota_grasp_opt}
\end{table}

\begin{figure}[htbp]
\centering
\noindent
\begin{minipage}[t]{0.48\textwidth}
    \vspace{0pt} 
    \centering
    \includegraphics[width=\linewidth]{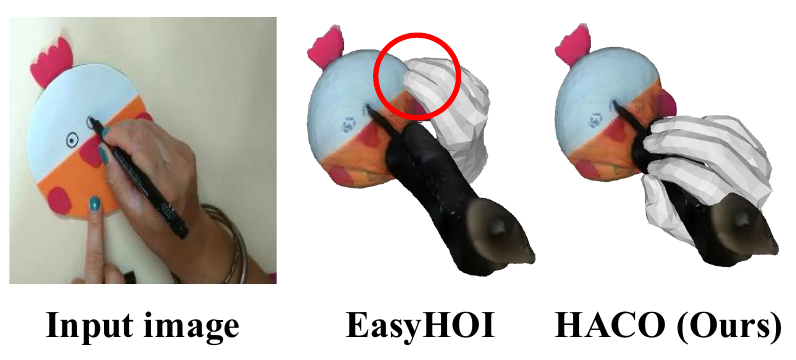}
    \vspace{-3mm}
    \captionof{figure}{\textbf{Qualitative comparison of 3D hand-object reconstruction on MOW~\cite{cao2021reconstructing} dataset.}}
    \label{fig:qual_ho_recon}
\end{minipage}%
\hfill
\begin{minipage}[t]{0.5\textwidth}
    \vspace{0pt} 
    \centering
    \scriptsize
    \setlength{\tabcolsep}{4.5pt}
    \captionof{table}{\textbf{Comparison of various contact estimation methods for 3D hand and object reconstruction with EasyHOI~\cite{liu2024easyhoi} on MOW~\cite{cao2021reconstructing} dataset.} PVE and PJE refer to MPVPE and MPJPE on vertices and joints between estimated and ground-truth 3D hand, respectively.}
    \begin{tabular}{lccccc}
        \toprule
        Methods & $\text{PVE}{\downarrow}$ & $\text{PJE}{\downarrow}$ & $\text{CD}_{\text{ho}}{\downarrow}$ & $\text{F-5}_{\text{ho}}{\uparrow}$ & $\text{F-10}_{\text{ho}}{\uparrow}$ \\
        \midrule
        EasyHOI~\cite{liu2024easyhoi} & 21.254 & 20.973 & 8.338 & 0.120 & 0.230 \\
        \midrule
        HACO~(Ours) & \textbf{21.093} & \textbf{20.845} & \textbf{8.186} & \textbf{0.122} & \textbf{0.231} \\
        \bottomrule
    \end{tabular}
    \label{tab:sota_hhor}
\end{minipage}
\vspace{-0.4cm}
\end{figure}

\noindent\textbf{3D hand and object reconstruction.}
Table~\ref{tab:sota_hhor} compares the performance of EasyHOI~\cite{liu2024easyhoi} between using its original contact estimation module and using HACO, keeping all other components unchanged.
This isolates the impact of the contact on reconstruction performance. HACO outperforms the original across all metrics, including MPVPE, MPJPE, Chamfer Distance~($\text{CD}_{\text{ho}}$), and F-scores.
These consistent gains suggest that HACO produces more accurate and physically plausible contact predictions, directly improving hand-object alignment and reconstruction quality.
Figure~\ref{fig:qual_ho_recon} further supports this, showing that the 3D reconstruction with contact from HACO more precisely captures interaction between the hand and the pen, unlike EasyHOI’s heuristic-based approach.

\section{Conclusion}
We propose HACO, a novel and powerful method that learns dense hand contact estimation from imbalanced data.
For class imbalance and spatial imbalance in hand contact datasets, we propose balanced contact sampling and vertex-level class-balanced loss.
As a result, our HACO outperforms previous methods by a significant margin on dense hand contact estimation and shows effectiveness on 3D grasp optimization and 3D hand and object reconstruction.

\section*{Acknowledgements}
This work was supported in part by the IITP grants [No.2021-0-01343, No.2023-0-00156, and Artificial Intelligence Graduate School Program (Seoul National University) - No.2021-0-02068], and the Industrial Technology Alchemist Project [No.RS-2024-00432410] funded by MOTIE, Korea.
\clearpage

\bibliography{references}{}

\begin{thebibliography}{10}

\bibitem{brahmbhatt2019contactdb}
Samarth Brahmbhatt, Cusuh Ham, Charles~C Kemp, and James Hays.
\newblock {ContactDB}: Analyzing and predicting grasp contact via thermal imaging.
\newblock In {\em CVPR}, 2019.

\bibitem{brown2020language}
Tom~B. Brown, Benjamin Mann, Nick Ryder, Melanie Subbiah, Jared Kaplan, Prafulla Dhariwal, Arvind Neelakantan, Pranav Shyam, Girish Sastry, Amanda Askell, Sandhini Agarwal, Ariel Herbert-Voss, Gretchen Krueger, Tom Henighan, Rewon Child, Aditya Ramesh, Daniel~M. Ziegler, Jeffrey Wu, Clemens Winter, Christopher Hesse, Mark Chen, Eric Sigler, Mateusz Litwin, Scott Gray, Benjamin Chess, Jack Clark, Christopher Berner, Sam McCandlish, Alec Radford, Ilya Sutskever, and Dario Amodei.
\newblock Language models are few-shot learners.
\newblock In {\em NeurIPS}, 2020.

\bibitem{cao2019learning}
Kaidi Cao, Colin Wei, Adrien Gaidon, Nikos Arechiga, and Tengyu Ma.
\newblock Learning imbalanced datasets with label-distribution-aware margin loss.
\newblock In {\em NeurIPS}, 2019.

\bibitem{cao2021reconstructing}
Zhe Cao, Ilija Radosavovic, Angjoo Kanazawa, and Jitendra Malik.
\newblock Reconstructing hand-object interactions in the wild.
\newblock In {\em ICCV}, 2021.

\bibitem{chao2021dexycb}
Yu-Wei Chao, Wei Yang, Yu~Xiang, Pavlo Molchanov, Ankur Handa, Jonathan Tremblay, Yashraj~S Narang, Karl Van~Wyk, Umar Iqbal, Stan Birchfield, Jan Kautz, and Dieter Fox.
\newblock {DexYCB}: A benchmark for capturing hand grasping of objects.
\newblock In {\em CVPR}, 2021.

\bibitem{cui2019class}
Yin Cui, Menglin Jia, Tsung-Yi Lin, Yang Song, and Serge Belongie.
\newblock Class-balanced loss based on effective number of samples.
\newblock In {\em CVPR}, 2019.

\bibitem{trimesh}
{Dawson-Haggerty et al.}
\newblock Trimesh.

\bibitem{deng2009imagenet}
Jia Deng, Wei Dong, Richard Socher, Li-Jia Li, Kai Li, and Li~Fei-Fei.
\newblock {ImageNet}: A large-scale hierarchical image database.
\newblock In {\em CVPR}, 2009.

\bibitem{devlin2019bert}
Jacob Devlin, Ming-Wei Chang, Kenton Lee, and Kristina Toutanova.
\newblock {BERT}: Pre-training of deep bidirectional {Transformers} for language understanding.
\newblock In {\em NAACL}, 2019.

\bibitem{dosovitskiy2020image}
Alexey Dosovitskiy, Lucas Beyer, Alexander Kolesnikov, Dirk Weissenborn, Xiaohua Zhai, Thomas Unterthiner, Mostafa Dehghani, Matthias Minderer, Georg Heigold, Sylvain Gelly, Jakob Uszkoreit, and Neil Houlsby.
\newblock An image is worth 16x16 words: {Transformers} for image recognition at scale.
\newblock In {\em ICLR}, 2021.

\bibitem{fan2021learning}
Zicong Fan, Adrian Spurr, Muhammed Kocabas, Siyu Tang, Michael~J Black, and Otmar Hilliges.
\newblock Learning to disambiguate strongly interacting hands via probabilistic per-pixel part segmentation.
\newblock In {\em 3DV}, 2021.

\bibitem{fan2023arctic}
Zicong Fan, Omid Taheri, Dimitrios Tzionas, Muhammed Kocabas, Manuel Kaufmann, Michael~J Black, and Otmar Hilliges.
\newblock {ARCTIC}: A dataset for dexterous bimanual hand-object manipulation.
\newblock In {\em CVPR}, 2023.

\bibitem{feng2021collaborative}
Yao Feng, Vasileios Choutas, Timo Bolkart, Dimitrios Tzionas, and Michael~J Black.
\newblock Collaborative regression of expressive bodies using moderation.
\newblock In {\em 3DV}, 2021.

\bibitem{grady2021contactopt}
Patrick Grady, Chengcheng Tang, Christopher~D Twigg, Minh Vo, Samarth Brahmbhatt, and Charles~C Kemp.
\newblock {ContactOpt}: Optimizing contact to improve grasps.
\newblock In {\em CVPR}, 2021.

\bibitem{hampali2020honnotate}
Shreyas Hampali, Mahdi Rad, Markus Oberweger, and Vincent Lepetit.
\newblock {HOnnotate}: A method for {3D} annotation of hand and object poses.
\newblock In {\em CVPR}, 2020.

\bibitem{hampali2022keypoint}
Shreyas Hampali, Sayan~Deb Sarkar, Mahdi Rad, and Vincent Lepetit.
\newblock {Keypoint Transformer}: Solving joint identification in challenging hands and object interactions for accurate {3D} pose estimation.
\newblock In {\em CVPR}, 2022.

\bibitem{hassan2019resolving}
Mohamed Hassan, Vasileios Choutas, Dimitrios Tzionas, and Michael~J Black.
\newblock Resolving {3D} human pose ambiguities with {3D} scene constraints.
\newblock In {\em ICCV}, 2019.

\bibitem{hassan2021populating}
Mohamed Hassan, Partha Ghosh, Joachim Tesch, Dimitrios Tzionas, and Michael~J Black.
\newblock Populating {3D} scenes by learning human-scene interaction.
\newblock In {\em CVPR}, 2021.

\bibitem{hasson2019learning}
Yana Hasson, Gul Varol, Dimitrios Tzionas, Igor Kalevatykh, Michael~J Black, Ivan Laptev, and Cordelia Schmid.
\newblock Learning joint reconstruction of hands and manipulated objects.
\newblock In {\em CVPR}, 2019.

\bibitem{he2016deep}
Kaiming He, Xiangyu Zhang, Shaoqing Ren, and Jian Sun.
\newblock Deep residual learning for image recognition.
\newblock In {\em CVPR}, 2016.

\bibitem{hu2024learning}
Junxing Hu, Hongwen Zhang, Zerui Chen, Mengcheng Li, Yunlong Wang, Yebin Liu, and Zhenan Sun.
\newblock Learning explicit contact for implicit reconstruction of hand-held objects from monocular images.
\newblock In {\em AAAI}, 2024.

\bibitem{huang2022capturing}
Chun-Hao~P Huang, Hongwei Yi, Markus H{\"o}schle, Matvey Safroshkin, Tsvetelina Alexiadis, Senya Polikovsky, Daniel Scharstein, and Michael~J Black.
\newblock Capturing and inferring dense full-body human-scene contact.
\newblock In {\em CVPR}, 2022.

\bibitem{kirillov2023segment}
Alexander Kirillov, Eric Mintun, Nikhila Ravi, Hanzi Mao, Chloe Rolland, Laura Gustafson, Tete Xiao, Spencer Whitehead, Alexander~C. Berg, Wan-Yen Lo, Piotr Doll{\'a}r, and Ross Girshick.
\newblock Segment {Anything}.
\newblock In {\em ICCV}, 2023.

\bibitem{kwon2021h2o}
Taein Kwon, Bugra Tekin, Jan St{\"u}hmer, Federica Bogo, and Marc Pollefeys.
\newblock {H2O}: Two hands manipulating objects for first person interaction recognition.
\newblock In {\em ICCV}, 2021.

\bibitem{lee2025semanticdraw}
Jaerin Lee, Daniel~Sungho Jung, Kanggeon Lee, and Kyoung~Mu Lee.
\newblock {SemanticDraw}: Towards real-time interactive content creation from image diffusion models.
\newblock In {\em CVPR}, 2025.

\bibitem{lengpolyloss}
Zhaoqi Leng, Mingxing Tan, Chenxi Liu, Ekin~Dogus Cubuk, Jay Shi, Shuyang Cheng, and Dragomir Anguelov.
\newblock {PolyLoss}: A polynomial expansion perspective of classification loss functions.
\newblock In {\em ICLR}, 2022.

\bibitem{li2025hybrik}
Jiefeng Li, Siyuan Bian, Chao Xu, Zhicun Chen, Lixin Yang, and Cewu Lu.
\newblock {HybrIK-X}: Hybrid analytical-neural inverse kinematics for whole-body mesh recovery.
\newblock {\em TPAMI}, 2025.

\bibitem{li2022cliff}
Zhihao Li, Jianzhuang Liu, Zhensong Zhang, Songcen Xu, and Youliang Yan.
\newblock {CLIFF}: Carrying location information in full frames into human pose and shape estimation.
\newblock In {\em ECCV}, 2022.

\bibitem{lin2021end}
Kevin Lin, Lijuan Wang, and Zicheng Liu.
\newblock End-to-end human pose and mesh reconstruction with {Transformers}.
\newblock In {\em CVPR}, 2021.

\bibitem{lin2021mesh}
Kevin Lin, Lijuan Wang, and Zicheng Liu.
\newblock Mesh graphormer.
\newblock In {\em ICCV}, 2021.

\bibitem{lin2017feature}
Tsung-Yi Lin, Piotr Doll{\'a}r, Ross Girshick, Kaiming He, Bharath Hariharan, and Serge Belongie.
\newblock Feature pyramid networks for object detection.
\newblock In {\em CVPR}, 2017.

\bibitem{lin2017focal}
Tsung-Yi Lin, Priya Goyal, Ross Girshick, Kaiming He, and Piotr Doll{\'a}r.
\newblock Focal loss for dense object detection.
\newblock In {\em ICCV}, 2017.

\bibitem{lin2023harmonious}
Zhifeng Lin, Changxing Ding, Huan Yao, Zengsheng Kuang, and Shaoli Huang.
\newblock Harmonious feature learning for interactive hand-object pose estimation.
\newblock In {\em CVPR}, 2023.

\bibitem{liu2023contactgen}
Shaowei Liu, Yang Zhou, Jimei Yang, Saurabh Gupta, and Shenlong Wang.
\newblock {ContactGen}: Generative contact modeling for grasp generation.
\newblock In {\em ICCV}, 2023.

\bibitem{liu2024easyhoi}
Yumeng Liu, Xiaoxiao Long, Zemin Yang, Yuan Liu, Marc Habermann, Christian Theobalt, Yuexin Ma, and Wenping Wang.
\newblock {EasyHOI}: Unleashing the power of large models for reconstructing hand-object interactions in the wild.
\newblock In {\em CVPR}, 2025.

\bibitem{liu2022hoi4d}
Yunze Liu, Yun Liu, Che Jiang, Kangbo Lyu, Weikang Wan, Hao Shen, Boqiang Liang, Zhoujie Fu, He~Wang, and Li~Yi.
\newblock {HOI4D}: A {4D} egocentric dataset for category-level human-object interaction.
\newblock In {\em CVPR}, 2022.

\bibitem{loper2015smpl}
Matthew Loper, Naureen Mahmood, Javier Romero, Gerard Pons-Moll, and Michael~J Black.
\newblock {SMPL}: A skinned multi-person linear model.
\newblock {\em ACM TOG}, 2015.

\bibitem{loshchilov2018decoupled}
Ilya Loshchilov and Frank Hutter.
\newblock Decoupled weight decay regularization.
\newblock In {\em ICLR}, 2019.

\bibitem{lu2025dposer}
Junzhe Lu, Jing Lin, Hongkun Dou, Ailing Zeng, Yue Deng, Xian Liu, Zhongang Cai, Lei Yang, Yulun Zhang, Haoqian Wang, and Ziwei Liu.
\newblock {DPoser-X}: Diffusion model as robust {3D} whole-body human pose prior.
\newblock In {\em ICCV}, 2025.

\bibitem{moon2023dataset}
Gyeongsik Moon, Shunsuke Saito, Weipeng Xu, Rohan Joshi, Julia Buffalini, Harley Bellan, Nicholas Rosen, Jesse Richardson, Mallorie Mize, Philippe de~Bree, Tomas Simon, Bo~Peng, Shubham Garg, Kevyn McPhail, and Takaaki Shiratori.
\newblock A dataset of relighted {3D} interacting hands.
\newblock In {\em NeurIPS}, 2023.

\bibitem{moon2020interhand2}
Gyeongsik Moon, Shoou-I Yu, He~Wen, Takaaki Shiratori, and Kyoung~Mu Lee.
\newblock {InterHand2.6M}: A dataset and baseline for {3D} interacting hand pose estimation from a single {RGB} image.
\newblock In {\em ECCV}, 2020.

\bibitem{muller2021self}
Lea Muller, Ahmed~AA Osman, Siyu Tang, Chun-Hao~P Huang, and Michael~J Black.
\newblock On self-contact and human pose.
\newblock In {\em CVPR}, 2021.

\bibitem{nam2024joint}
Hyeongjin Nam, Daniel~Sungho Jung, Gyeongsik Moon, and Kyoung~Mu Lee.
\newblock Joint reconstruction of {3D} human and object via contact-based refinement {Transformer}.
\newblock In {\em CVPR}, 2024.

\bibitem{nam2023cyclic}
Hyeongjin Nam, Daniel~Sungho Jung, Yeonguk Oh, and Kyoung~Mu Lee.
\newblock Cyclic test-time adaptation on monocular video for {3D} human mesh reconstruction.
\newblock In {\em ICCV}, 2023.

\bibitem{park2023extract}
JoonKyu Park, Daniel~Sungho Jung, Gyeongsik Moon, and Kyoung~Mu Lee.
\newblock Extract-and-adaptation network for {3D} interacting hand mesh recovery.
\newblock In {\em ICCV}, 2023.

\bibitem{park2022handoccnet}
JoonKyu Park, Yeonguk Oh, Gyeongsik Moon, Hongsuk Choi, and Kyoung~Mu Lee.
\newblock {HandOccNet}: Occlusion-robust {3D} hand mesh estimation network.
\newblock In {\em CVPR}, 2022.

\bibitem{paszkepytorch}
Adam Paszke, Sam Gross, Francisco Massa, Adam Lerer, James Bradbury, Gregory Chanan, Trevor Killeen, Zeming Lin, Natalia Gimelshein, Luca Antiga, Alban Desmaison, Andreas Köpf, Edward Yang, Zach DeVito, Martin Raison, Alykhan Tejani, Sasank Chilamkurthy, Benoit Steiner, Lu~Fang, Junjie Bai, and Soumith Chintala.
\newblock {PyTorch}: An imperative style, high-performance deep learning library.
\newblock In {\em NeurIPS}, 2019.

\bibitem{pavlakos2019expressive}
Georgios Pavlakos, Vasileios Choutas, Nima Ghorbani, Timo Bolkart, Ahmed~AA Osman, Dimitrios Tzionas, and Michael~J Black.
\newblock Expressive body capture: {3D} hands, face, and body from a single image.
\newblock In {\em CVPR}, 2019.

\bibitem{pavlakos2024reconstructing}
Georgios Pavlakos, Dandan Shan, Ilija Radosavovic, Angjoo Kanazawa, David Fouhey, and Jitendra Malik.
\newblock Reconstructing hands in {3D} with {Transformers}.
\newblock In {\em CVPR}, 2024.

\bibitem{prakash2025synthesizing}
Aditya Prakash, Benjamin Lundell, Dmitry Andreychuk, David Forsyth, Saurabh Gupta, and Harpreet Sawhney.
\newblock How do i do that? {Synthesizing} {3D} hand motion and contacts for everyday interactions.
\newblock In {\em CVPR}, 2025.

\bibitem{ravi2024sam}
Nikhila Ravi, Valentin Gabeur, Yuan-Ting Hu, Ronghang Hu, Chaitanya Ryali, Tengyu Ma, Haitham Khedr, Roman Rädle, Chloe Rolland, Laura Gustafson, Eric Mintun, Junting Pan, Kalyan~Vasudev Alwala, Nicolas Carion, Chao-Yuan Wu, Ross Girshick, Piotr Dollár, and Christoph Feichtenhofer.
\newblock {SAM 2}: {Segment Anything} in images and videos.
\newblock In {\em ICLR}, 2025.

\bibitem{ridnik2021asymmetric}
Tal Ridnik, Emanuel Ben-Baruch, Nadav Zamir, Asaf Noy, Itamar Friedman, Matan Protter, and Lihi Zelnik-Manor.
\newblock Asymmetric loss for multi-label classification.
\newblock In {\em ICCV}, 2021.

\bibitem{rombach2022high}
Robin Rombach, Andreas Blattmann, Dominik Lorenz, Patrick Esser, and Bj{\"o}rn Ommer.
\newblock High-resolution image synthesis with latent diffusion models.
\newblock In {\em CVPR}, 2022.

\bibitem{romero2017embodied}
Javier Romero, Dimitrios Tzionas, and Michael~J Black.
\newblock Embodied hands: Modeling and capturing hands and bodies together.
\newblock {\em ACM TOG}, 2017.

\bibitem{shimada2023decaf}
Soshi Shimada, Vladislav Golyanik, Patrick P{\'e}rez, and Christian Theobalt.
\newblock Decaf: Monocular deformation capture for face and hand interactions.
\newblock {\em ACM TOG}, 2023.

\bibitem{sohn2015learning}
Kihyuk Sohn, Honglak Lee, and Xinchen Yan.
\newblock Learning structured output representation using deep conditional generative models.
\newblock In {\em NeurIPS}, 2015.

\bibitem{sun2019deep}
Ke~Sun, Bin Xiao, Dong Liu, and Jingdong Wang.
\newblock Deep high-resolution representation learning for human pose estimation.
\newblock In {\em CVPR}, 2019.

\bibitem{tripathi2023deco}
Shashank Tripathi, Agniv Chatterjee, Jean-Claude Passy, Hongwei Yi, Dimitrios Tzionas, and Michael~J Black.
\newblock {DECO}: Dense estimation of {3D} human-scene contact in the wild.
\newblock In {\em ICCV}, 2023.

\bibitem{tzionas2016capturing}
Dimitrios Tzionas, Luca Ballan, Abhilash Srikantha, Pablo Aponte, Marc Pollefeys, and Juergen Gall.
\newblock Capturing hands in action using discriminative salient points and physics simulation.
\newblock {\em IJCV}, 2016.

\bibitem{vaswani2017attention}
Ashish Vaswani, Noam Shazeer, Niki Parmar, Jakob Uszkoreit, Llion Jones, Aidan~N Gomez, {\L}ukasz Kaiser, and Illia Polosukhin.
\newblock Attention is all you need.
\newblock In {\em NeurIPS}, 2017.

\bibitem{wu2024dice}
Qingxuan Wu, Zhiyang Dou, Sirui Xu, Soshi Shimada, Chen Wang, Zhengming Yu, Yuan Liu, Cheng Lin, Zeyu Cao, Taku Komura, Vladislav Golyanik, Christian Theobalt, Wenping Wang, and Lingjie Liu.
\newblock {DICE}: End-to-end deformation capture of hand-face interactions from a single image.
\newblock In {\em ICLR}, 2025.

\bibitem{xie2024rhobin}
Xianghui Xie, Xi~Wang, Nikos Athanasiou, Bharat~Lal Bhatnagar, Chun-Hao~P. Huang, Kaichun Mo, Hao Chen, Xia Jia, Zerui Zhang, Liangxian Cui, Xiao Lin, Bingqiao Qian, Jie Xiao, Wenfei Yang, Hyeongjin Nam, Daniel~Sungho Jung, Kihoon Kim, Kyoung~Mu Lee, Otmar Hilliges, and Gerard Pons-Moll.
\newblock {RHOBIN Challenge}: Reconstruction of human object interaction.
\newblock {\em arXiv preprint arXiv:2401.04143}, 2024.

\bibitem{xu2022vitpose}
Yufei Xu, Jing Zhang, Qiming Zhang, and Dacheng Tao.
\newblock {ViTPose}: Simple {Vision Transformer} baselines for human pose estimation.
\newblock In {\em NeurIPS}, 2022.

\bibitem{yang2024depth}
Lihe Yang, Bingyi Kang, Zilong Huang, Xiaogang Xu, Jiashi Feng, and Hengshuang Zhao.
\newblock {Depth Anything}: Unleashing the power of large-scale unlabeled data.
\newblock In {\em CVPR}, 2024.

\bibitem{yin2023hi4d}
Yifei Yin, Chen Guo, Manuel Kaufmann, Juan~Jose Zarate, Jie Song, and Otmar Hilliges.
\newblock {Hi4D}: {4D} instance segmentation of close human interaction.
\newblock In {\em CVPR}, 2023.

\bibitem{zhang2023pymaf}
Hongwen Zhang, Yating Tian, Yuxiang Zhang, Mengcheng Li, Liang An, Zhenan Sun, and Yebin Liu.
\newblock {PyMAF-X}: Towards well-aligned full-body model regression from monocular images.
\newblock {\em TPAMI}, 2023.

\bibitem{zhao2023poseformerv2}
Qitao Zhao, Ce~Zheng, Mengyuan Liu, Pichao Wang, and Chen Chen.
\newblock {PoseFormerV2}: Exploring frequency domain for efficient and robust {3D} human pose estimation.
\newblock In {\em CVPR}, 2023.

\end{thebibliography}
\bibliographystyle{plain}
\clearpage

\section*{NeurIPS Paper Checklist}

The checklist is designed to encourage best practices for responsible machine learning research, addressing issues of reproducibility, transparency, research ethics, and societal impact. Do not remove the checklist: {\bf The papers not including the checklist will be desk rejected.} The checklist should follow the references and follow the (optional) supplemental material.  The checklist does NOT count towards the page
limit. 

Please read the checklist guidelines carefully for information on how to answer these questions. For each question in the checklist:
\begin{itemize}
    \item You should answer \answerYes{}, \answerNo{}, or \answerNA{}.
    \item \answerNA{} means either that the question is Not Applicable for that particular paper or the relevant information is Not Available.
    \item Please provide a short (1–2 sentence) justification right after your answer (even for NA). 
\end{itemize}

{\bf The checklist answers are an integral part of your paper submission.} They are visible to the reviewers, area chairs, senior area chairs, and ethics reviewers. You will be asked to also include it (after eventual revisions) with the final version of your paper, and its final version will be published with the paper.

The reviewers of your paper will be asked to use the checklist as one of the factors in their evaluation. While "\answerYes{}" is generally preferable to "\answerNo{}", it is perfectly acceptable to answer "\answerNo{}" provided a proper justification is given (e.g., "error bars are not reported because it would be too computationally expensive" or "we were unable to find the license for the dataset we used"). In general, answering "\answerNo{}" or "\answerNA{}" is not grounds for rejection. While the questions are phrased in a binary way, we acknowledge that the true answer is often more nuanced, so please just use your best judgment and write a justification to elaborate. All supporting evidence can appear either in the main paper or the supplemental material, provided in appendix. If you answer \answerYes{} to a question, in the justification please point to the section(s) where related material for the question can be found.

IMPORTANT, please:
\begin{itemize}
    \item {\bf Delete this instruction block, but keep the section heading ``NeurIPS Paper Checklist"},
    \item  {\bf Keep the checklist subsection headings, questions/answers and guidelines below.}
    \item {\bf Do not modify the questions and only use the provided macros for your answers}.
\end{itemize}


\begin{enumerate}

\item {\bf Claims}
    \item[] Question: Do the main claims made in the abstract and introduction accurately reflect the paper's contributions and scope?
    \item[] Answer: \answerYes{} 
    \item[] Justification: We clearly state the core motivation to understand dense hand contact, outlines the challenges of class and spatial imbalance, introduces the proposed methods including balanced contact sampling and vertex-level class-balanced loss, and presents the contribution of a framework for learning dense hand contact estimation from imbalanced data. These claims are addressed in the introduction and supported by experimental results, accurately reflecting the paper's contributions and scope.

\item {\bf Limitations}
    \item[] Question: Does the paper discuss the limitations of the work performed by the authors?
    \item[] Answer: \answerYes{} 
    \item[] Justification: The limitations are discussed in the appendix, where we address the scope and boundaries of our method, including future directions.

\item {\bf Theory assumptions and proofs}
    \item[] Question: For each theoretical result, does the paper provide the full set of assumptions and a complete (and correct) proof?
    \item[] Answer: \answerNA{} 
    \item[] Justification: Our paper does not contain any theoretical result.

    \item {\bf Experimental result reproducibility}
    \item[] Question: Does the paper fully disclose all the information needed to reproduce the main experimental results of the paper to the extent that it affects the main claims and/or conclusions of the paper (regardless of whether the code and data are provided or not)?
    \item[] Answer: \answerYes{} 
    \item[] Justification: The paper provides all necessary details to reproduce the main experiments, including data preprocessing, model architecture, training settings, and evaluation protocols.

\item {\bf Open access to data and code}
    \item[] Question: Does the paper provide open access to the data and code, with sufficient instructions to faithfully reproduce the main experimental results, as described in supplemental material?
    \item[] Answer: \answerYes{} 
    \item[] Justification: In the abstract of the paper, we provide URL to our publicly available code in an open repository.

\item {\bf Experimental setting/details}
    \item[] Question: Does the paper specify all the training and test details (e.g., data splits, hyperparameters, how they were chosen, type of optimizer, etc.) necessary to understand the results?
    \item[] Answer: \answerYes{} 
    \item[] Justification: The paper specifies dataset splits, model architecture, training schedules, optimizer settings, and other key hyperparameters necessary to understand and reproduce the results.

\item {\bf Experiment statistical significance}
    \item[] Question: Does the paper report error bars suitably and correctly defined or other appropriate information about the statistical significance of the experiments?
    \item[] Answer: \answerNo{} 
    \item[] Justification: The results are obtained from standard randomly seeded runs without cherry-picking, offering a fair and representative evaluation despite not including error bars.

\item {\bf Experiments compute resources}
    \item[] Question: For each experiment, does the paper provide sufficient information on the computer resources (type of compute workers, memory, time of execution) needed to reproduce the experiments?
    \item[] Answer: \answerYes{} 
    \item[] Justification: The paper specifies the type of GPU used, batch size, number of training epochs, and key runtime settings needed to reproduce the experiments.
    
\item {\bf Code of ethics}
    \item[] Question: Does the research conducted in the paper conform, in every respect, with the NeurIPS Code of Ethics \url{https://neurips.cc/public/EthicsGuidelines}?
    \item[] Answer: \answerYes{} 
    \item[] Justification: The research adheres to the NeurIPS Code of Ethics, with no use of sensitive data or methods posing foreseeable ethical risks.

\item {\bf Broader impacts}
    \item[] Question: Does the paper discuss both potential positive societal impacts and negative societal impacts of the work performed?
    \item[] Answer: \answerYes{} 
    \item[] Justification: The paper discusses societal impacts in the appendix, including both potential benefits and risks.
    
\item {\bf Safeguards}
    \item[] Question: Does the paper describe safeguards that have been put in place for responsible release of data or models that have a high risk for misuse (e.g., pretrained language models, image generators, or scraped datasets)?
    \item[] Answer: \answerNA{} 
    \item[] Justification: The work does not involve releasing models or data with high risk of misuse, so specific safeguards are not applicable.

\item {\bf Licenses for existing assets}
    \item[] Question: Are the creators or original owners of assets (e.g., code, data, models), used in the paper, properly credited and are the license and terms of use explicitly mentioned and properly respected?
    \item[] Answer: \answerYes{} 
    \item[] Justification: All external datasets and code-bases used in the paper are properly cited.

\item {\bf New assets}
    \item[] Question: Are new assets introduced in the paper well documented and is the documentation provided alongside the assets?
    \item[] Answer: \answerNA{} 
    \item[] Justification: The paper does not introduce new assets such as datasets that require separate documentation.

\item {\bf Crowdsourcing and research with human subjects}
    \item[] Question: For crowdsourcing experiments and research with human subjects, does the paper include the full text of instructions given to participants and screenshots, if applicable, as well as details about compensation (if any)? 
    \item[] Answer: \answerNA{} 
    \item[] Justification: We mainly utilize publicly available datasets for our experiments.

\item {\bf Institutional review board (IRB) approvals or equivalent for research with human subjects}
    \item[] Question: Does the paper describe potential risks incurred by study participants, whether such risks were disclosed to the subjects, and whether Institutional Review Board (IRB) approvals (or an equivalent approval/review based on the requirements of your country or institution) were obtained?
    \item[] Answer: \answerNA{} 
    \item[] Justification: We mainly utilize publicly available datasets for our experiments.

\item {\bf Declaration of LLM usage}
    \item[] Question: Does the paper describe the usage of LLMs if it is an important, original, or non-standard component of the core methods in this research? Note that if the LLM is used only for writing, editing, or formatting purposes and does not impact the core methodology, scientific rigorousness, or originality of the research, declaration is not required.
    \item[] Answer: \answerNA{} 
    \item[] Justification: LLM is not used in any critical component of the research process for this paper.

\end{enumerate}
\clearpage

\appendix
\section{Appendix}\label{sec:appendix}
  
\setcounter{table}{0}
\setcounter{figure}{0}
\renewcommand{\thetable}{A\arabic{table}}   
\renewcommand{\thefigure}{A\arabic{figure}}

In this appendix, we provide additional technical details and experimental results that were omitted from the main manuscript due to space constraints.
The contents are summarized below:
\begin{itemize}
\item \ref{sec:loss_functions}. Details of loss functions
\item \ref{sec:contact_init}. Details of contact initialization
\item \ref{sec:gt_contact}. Details of dense hand contact labels
\item \ref{sec:class_imbalance}. More examples of class imbalance issue
\item \ref{sec:spatial_imbalance}. More examples of spatial imbalance issue
\item \ref{sec:more_sota_comparison}. More comparisons with state-of-the-art methods
\item \ref{sec:more_abl_study}. More ablation studies
\item \ref{sec:more_backbones}. Quantitative results with different backbones
\item \ref{sec:computation}. Computational requirements
\item \ref{sec:more_qualitative}. More qualitative results
\item \ref{sec:limitations}. Limitations and societal impacts
\end{itemize}

\subsection{Details of loss functions}
\label{sec:loss_functions}

\noindent\textbf{Smoothness loss.}
We penalize spatially isolated or fragmented contact predictions by leveraging the mesh topology of the hand. Let $p_v \in [0,1]$ denote the predicted contact probability after the sigmoid for each vertex $v \in V$, and let $\mathbf{A} \in \{0,1\}^{V \times V}$ be a sparse adjacency matrix representing vertex connectivity in the mesh. 
Here, $A_{vu} = 1$ indicates that vertex $u$ is connected to vertex $v$, and $0$ otherwise.
For numerical stability, we also include self-connections so that each vertex is connected to itself.
For each vertex $v \in V$, we compute the aggregated contact prediction over its neighbors as:
\begin{equation}
\hat{p}_v = \sum_{u \in V} A_{vu} \cdot p_u,
\end{equation}
and similarly compute the aggregated non-contact prediction as:
\begin{equation}
\hat{q}_v = \sum_{u \in V} A_{vu} \cdot (1 - p_u),
\end{equation}
where $\hat{p}_v$ and $\hat{q}_v$ represent the neighborhood-smoothed estimates for contact and non-contact probabilities, respectively.
To maintain consistent scale across batches, we further normalize the aggregated contact and non-contact maps by their maximum values within each batch before computing the isolation score.
To assess spatial inconsistency, we compute the discrepancy between each vertex's prediction and its neighborhood average. Specifically, we define the isolation score as:
\begin{equation}
s_v = \left| p_v - \hat{p}_v \right| + \left| (1 - p_v) - \hat{q}_v \right|.
\end{equation}
This term penalizes contact predictions that sharply differ from those of their neighbors, whether in the contact or non-contact region.
To prevent the isolation score from disproportionately penalizing spatially large or confident regions, we normalize the total discrepancy by the sum of neighborhood weights:
\begin{equation}
n_v = \sum_{u \in V} A_{vu} = \hat{p}_v + \hat{q}_v,
\end{equation}
which corresponds to the degree of vertex $v$ in the mesh. 
This normalization accounts for how many vertices contribute to the isolation score at each location, ensuring that the loss measures average inconsistency rather than accumulating errors over larger neighborhoods. 
As a result, the model is encouraged to produce smooth contact maps without being biased against large spatially coherent contact regions.
We then define the final smoothness regularization loss as:
\begin{equation}
\mathcal{L}_{\text{iso}} = \log\left(1 + \frac{ \sum_{v \in V} s_v }{ \sum_{v \in V} n_v + \epsilon } \right),
\end{equation}
where $\epsilon$ is a small constant added for numerical stability. 
This loss encourages smooth transitions and penalizes sharp discontinuities in the prediction map. 
By suppressing highly localized contact signals and promoting coherent spatial clusters, it guides the model to favor fewer but larger contact regions, better reflecting real-world hand interaction patterns.

\noindent\textbf{Regularization loss.}
To prevent the model from predicting arbitrarily skewed contact distributions, we introduce a global regularization loss that encourages predictions to stay close to the dataset-wide mean contact pattern. 
Let $p_v \in [0,1]$ denote the predicted contact probability after the sigmoid for each vertex $v \in V$, and let $\bar{p}_v \in [0,1]$ denote the dataset-wide average contact probability at vertex $v$, computed over all training samples. We then define the regularization loss as:
\begin{equation}
\mathcal{L}_{\text{reg}} = \frac{1}{|V|} \sum_{v \in V} \left| p_v - \bar{p}_v \right|,
\end{equation}
which is an L1 loss between the predicted contact probabilities and the mean distribution. 
This encourages the model to avoid degenerate or overly confident predictions that deviate significantly from typical contact patterns, providing a soft global constraint during training.

\subsection{Details of contact initialization}
\label{sec:contact_init}
To stabilize the training of HACO on dense hand contact estimation, we propose a contact initialization technique that utilizes a learnable embedding to learn the most effective initial contact for dense hand contact estimation.
In Table~\ref{tab:supp_abl_init_contact}, our proposed contact initialization outperforms all other initialization methods, achieving the superior performance in precision, recall, and F1-score. 
Unlike other approaches that rely on fixed priors or simplistic assumptions, our method introduces a contact representation learned from input data, enabling more accurate and robust initialization during inference. 
This formulation captures the most useful initial prediction compared to static or handcrafted methods, leading to gains across all metrics. 
Such consistent improvement demonstrates that learning contact directly from large-scale datasets, rather than relying on pre-defined numbers or dataset averages, is crucial for precise contact estimation.

\begin{table}[htbp]
\centering
\small
\setlength{\tabcolsep}{4pt}  
\caption{\textbf{Comparison of various contact initialization on MOW~\cite{cao2021reconstructing} dataset.}}
\begin{tabular}{lcccc} \toprule
Methods &  $\text{Precision}{\uparrow}$ & $\text{Recall}{\uparrow}$ & $\text{F1-Score}{\uparrow}$ \\ 
\midrule
No initialization & 0.508 & 0.589 & 0.503 \\
No contact & 0.514 & 0.532 & 0.482 \\ 
Full contact & 0.511 & 0.551 & 0.492 \\ 
Mean of DexYCB~\cite{chao2021dexycb} & 0.520 & 0.588 & 0.512 \\ 
Mean of MOW~\cite{cao2021reconstructing} & 0.511 & \underline{0.600} & \underline{0.515} \\ 
Mean of IH26M~\cite{moon2020interhand2} & \underline{0.522} & 0.567 & 0.505 \\ 
Ours & \textbf{0.525} & \textbf{0.607} & \textbf{0.522} \\ 
\bottomrule
\end{tabular}
\label{tab:supp_abl_init_contact}
\end{table}

\subsection{Details of dense hand contact labels}
\label{sec:gt_contact}
Following the previous works on dense human-scene contact estimation~\cite{hassan2021populating, huang2022capturing, nam2024joint}, we employ distance-based thresholding to generate ground-truth dense hand contact labels. This is implemented using the Trimesh library~\cite{trimesh}. Given a ground-truth hand mesh and the mesh of the interacting entity (i.e., an object for hand-object interaction, another hand for hand-hand interaction, a face for hand-face interaction, the environment for hand-scene interaction, or the body for hand-body interaction), we construct a proximity query from the interacting mesh and compute the closest surface point on it for each vertex of the hand mesh. Contact is then determined by thresholding the Euclidean distance between each hand vertex and its corresponding closest point. The resulting contact labels are binary values per vertex, indicating whether each vertex is in contact.

To set these thresholds, we manually inspected the ground-truth meshes and the resulting contact labels produced under fixed threshold values. 
We observed that the accuracy of 3D annotations varies significantly across datasets, making a single threshold unsuitable for all cases. 
A uniform threshold often led to incorrect labeling and semantically implausible contact regions.
Based on our analysis, we set the contact distance threshold to 1 cm for the following datasets: ObMan~\cite{hasson2019learning}, DexYCB~\cite{chao2021dexycb}, HO3D~\cite{hampali2020honnotate}, H2O3D~\cite{hampali2022keypoint}, ARCTIC~\cite{fan2023arctic}, HOI4D~\cite{liu2022hoi4d}, H2O~\cite{kwon2021h2o}, and PROX~\cite{hassan2019resolving}. 
For the MOW~\cite{cao2021reconstructing} dataset, we used a threshold of 3.5 cm due to its lower mesh fidelity. 
For InterHand2.6M~\cite{moon2020interhand2} and HIC~\cite{tzionas2016capturing}, we adopted a smaller threshold of 0.5 cm to better capture fine-grained hand-hand interactions.

Although most datasets produced reasonable contact labels with a 1 cm threshold, MOW required a larger value of 3.5 cm due to coarse mesh annotations generated through optimization on in-the-wild images. For hand-hand interaction datasets, the 1 cm threshold, which is approximately the width of a finger, often resulted in inaccurate and semantically implausible contact labels. To address this, we adopted a smaller threshold of 0.5 cm. For datasets with existing annotated contact labels, such as RICH~\cite{huang2022capturing}, Decaf~\cite{shimada2023decaf}, and Hi4D~\cite{yin2023hi4d}, we directly used the provided ground-truth annotations.

\begin{figure*}[t]
\begin{center}
\includegraphics[width=1.0\linewidth]{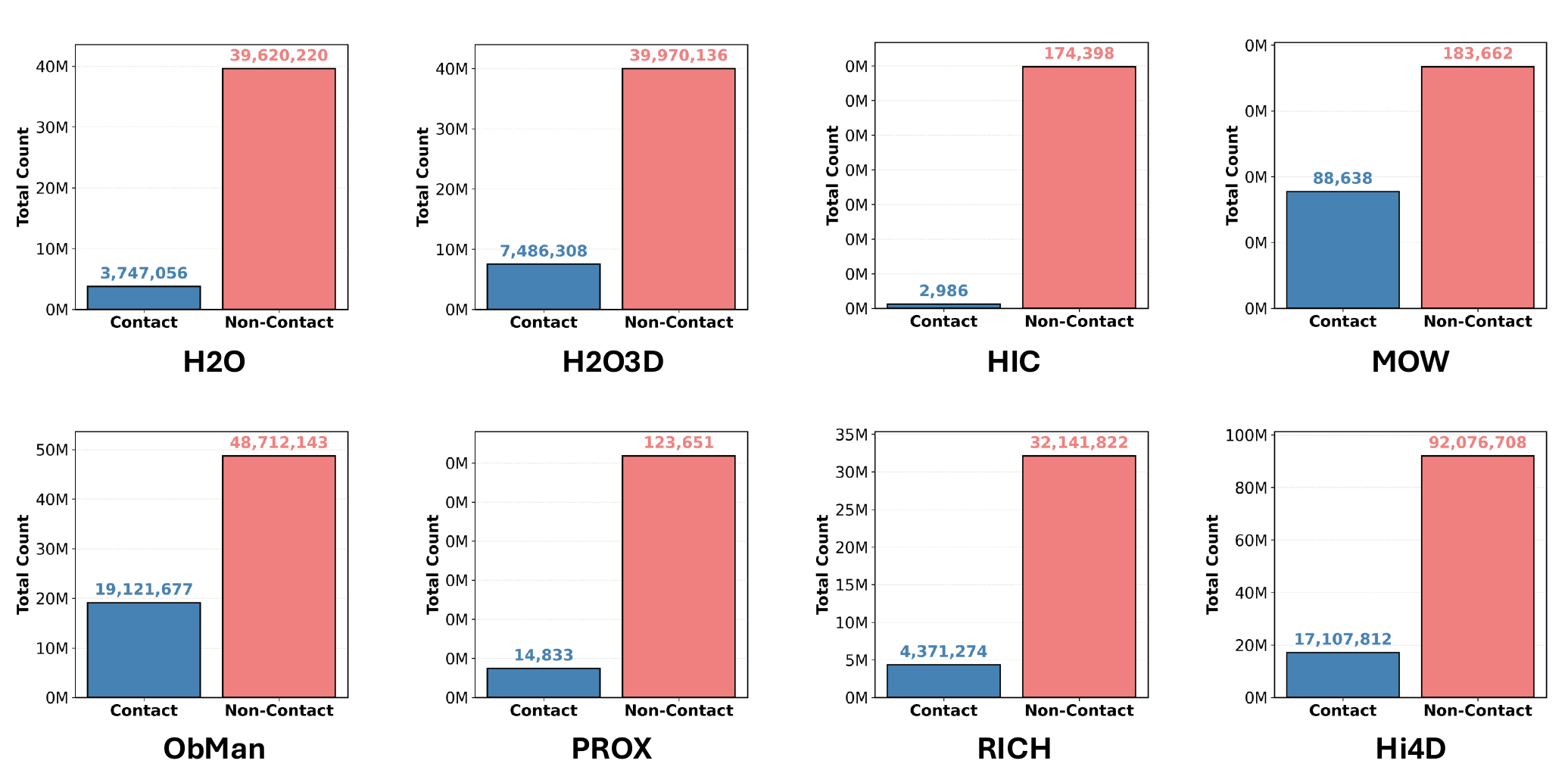}
\end{center}
\caption{
\textbf{More examples of class imbalance between contact and non-contact on H2O~\cite{kwon2021h2o}, H2O3D~\cite{hampali2022keypoint}, HIC~\cite{tzionas2016capturing}, MOW~\cite{cao2021reconstructing}, ObMan~\cite{hasson2019learning}, PROX~\cite{hassan2019resolving}, RICH~\cite{huang2022capturing}, Hi4D~\cite{yin2023hi4d}.}}
\label{fig:supp_class_imbalance}
\end{figure*}

\subsection{More examples of class imbalance issue}
\label{sec:class_imbalance}
Figure~\ref{fig:supp_class_imbalance} presents the class imbalance issue across datasets~\cite{kwon2021h2o, hampali2022keypoint, tzionas2016capturing, cao2021reconstructing, hasson2019learning, hassan2019resolving, huang2022capturing, yin2023hi4d} omitted from the main paper due to space constraints. 
In general, all datasets exhibit a severe class imbalance, with the majority of vertices corresponding to non-contact. 
Notably, some datasets such as H2O and HIC display highly skewed ratios of 10.57:1 and 58.41:1, respectively, between non-contact and contact vertices.
It is important to note that contact and non-contact counts are computed at the vertex-level rather than per-hand instance. 
Since the ground-truth contact labels are provided per-hand, rather than per-vertex for each image, sampling must be performed at the hand-level.
This makes it challenging to tackle class imbalance issue in per-hand manner as the class imbalance issue is present in per-vertex.
Hence, it is essential to design a principled, hand-level proxy that reflects the underlying per-vertex imbalance.
To tackle this challenge, we introduce balanced contact sampling (BCS), which constructs multiple sampling groups in hand-level but still fairly represent both contact and non-contact in per-vertex level.
Please refer to the main paper for further details on the BCS.

\subsection{More examples of spatial imbalance issue}
\label{sec:spatial_imbalance}
\begin{figure*}[t]
\begin{center}
\includegraphics[width=1.0\linewidth]{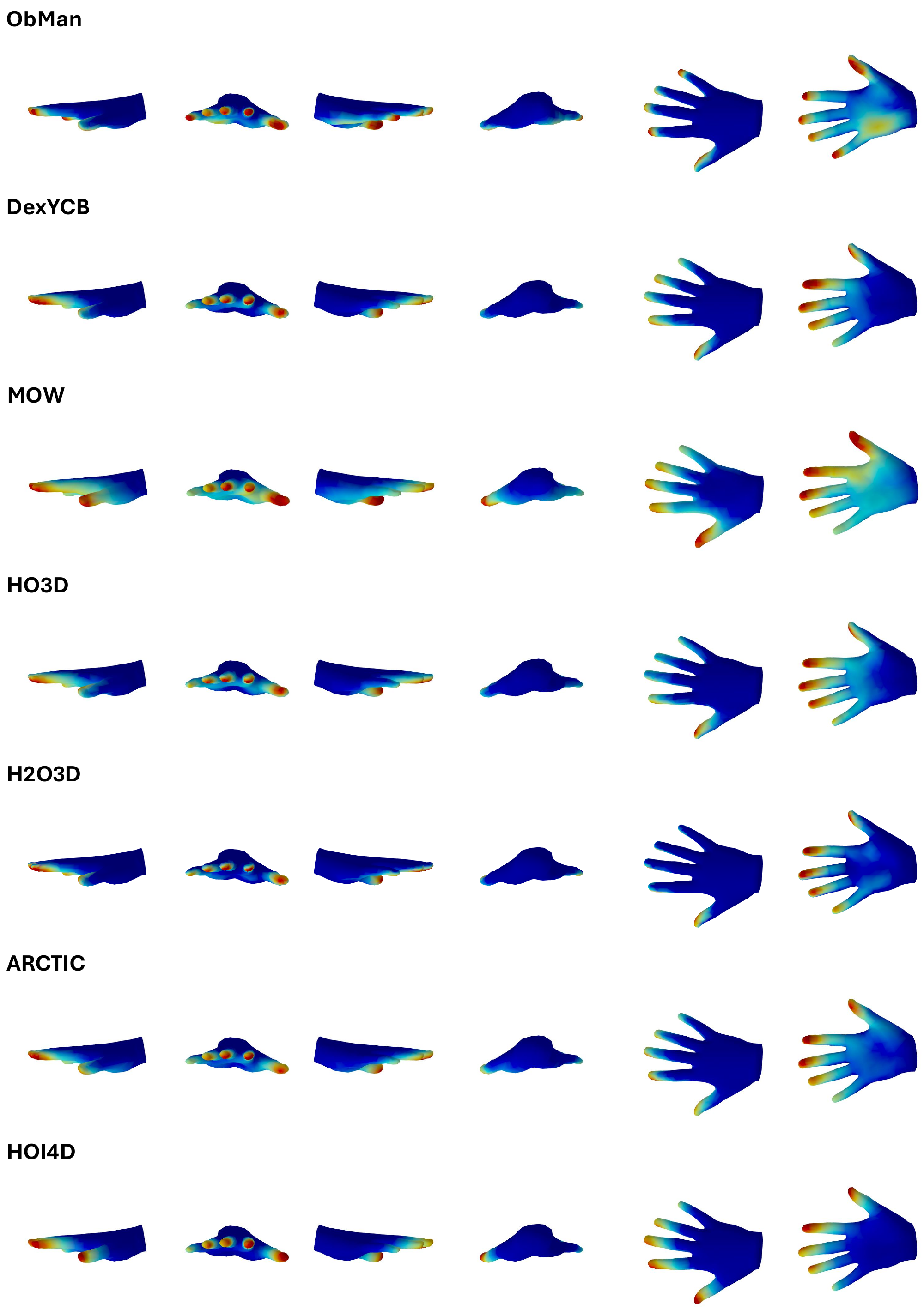}
\end{center}
\caption{
\textbf{Dataset-wise dense hand contact mean.} These heatmaps show average hand contact of ground-truths from ObMan~\cite{hasson2019learning}, DexYCB~\cite{chao2021dexycb}, MOW~\cite{cao2021reconstructing}, HO3D~\cite{hampali2020honnotate}, H2O3D~\cite{hampali2022keypoint}, ARCTIC~\cite{fan2023arctic}, HOI4D~\cite{liu2022hoi4d} dataset.}
\label{fig:hand_contact_mean}
\end{figure*}

\begin{figure*}[t]
\begin{center}
\includegraphics[width=1.0\linewidth]{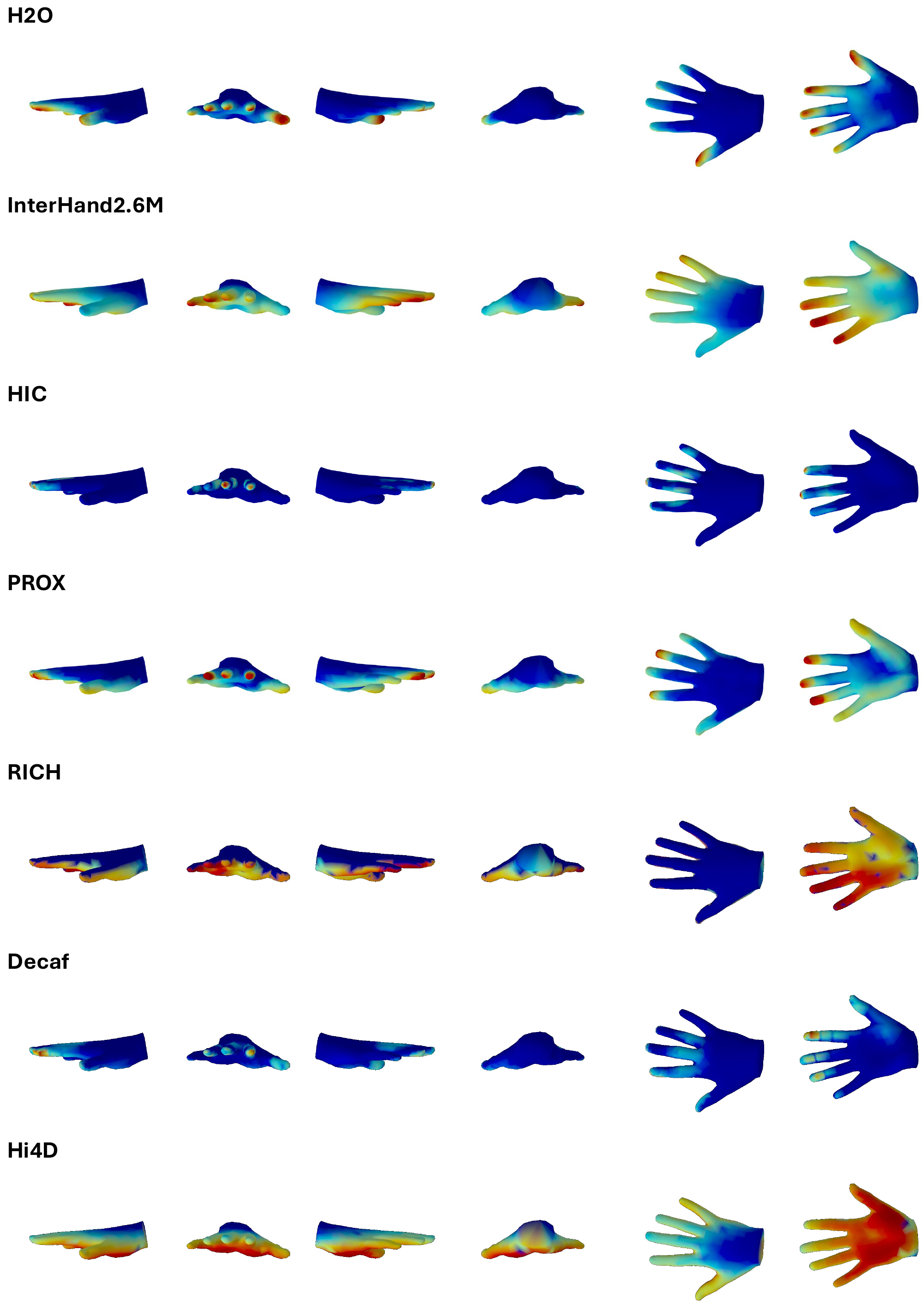}
\end{center}
\caption{
\textbf{Dataset-wise dense hand contact mean.} These heatmaps show average hand contact of ground-truths from H2O~\cite{kwon2021h2o}, InterHand2.6M~\cite{moon2020interhand2}, HIC~\cite{tzionas2016capturing}, PROX~\cite{hassan2019resolving}, RICH~\cite{huang2022capturing}, Decaf~\cite{shimada2023decaf}, Hi4D~\cite{yin2023hi4d} dataset.}
\label{fig:hand_contact_mean_more}
\end{figure*}

We visualize the dataset-wide mean contact maps for all dense hand contact datasets used in training HACO in Figure~\ref{fig:hand_contact_mean} and Figure~\ref{fig:hand_contact_mean_more}. Among hand-object interaction datasets, DexYCB~\cite{chao2021dexycb}, HO3D~\cite{hampali2020honnotate}, H2O3D~\cite{hampali2022keypoint}, ARCTIC~\cite{fan2023arctic}, HOI4D~\cite{liu2022hoi4d}, and H2O~\cite{kwon2021h2o} exhibit significant spatial imbalance, with ground-truth contact concentrated predominantly on the fingertips.
ObMan~\cite{hasson2019learning} and MOW~\cite{cao2021reconstructing} display more promising contact distributions, including high contact probability on the hypothenar region. However, ObMan is a synthetic dataset with domain gap towards real images, and MOW, while showing diverse palmar contact patterns, lacks dorsal contact and contains fewer than 1,000 samples, making it insufficient for large-scale training.
Among hand-hand interaction datasets, HIC~\cite{tzionas2016capturing} shows severe spatial imbalance, with contact limited to small fingertip regions and regions between fingers. In contrast, InterHand2.6M~\cite{moon2020interhand2} offers a more favorable contact distribution, covering nearly all hand regions except for a small area on the dorsum.
Hand-scene interaction datasets~\cite{hassan2019resolving, huang2022capturing} tend to involve large contact areas, primarily on the palmar side. However, interactions are often limited to flat surfaces such as ground or walls, resulting in low diversity of contact types despite broad coverage. 
The hand-face interaction dataset, Decaf~\cite{shimada2023decaf}, also exhibits poor spatial diversity due to its limited action types, such as poking or touching the chin. 
For the hand-body interaction dataset, Hi4D~\cite{yin2023hi4d} includes extensive contact regions, but most interactions involve full palmar contact during actions such as hugging or patting.
As visualized in the dataset-wide mean contact heatmaps, spatial imbalance is a prevalent issue across existing dense hand contact datasets. To address this, we propose the vertex-level class-balanced (VCB) loss, which mitigates spatial imbalance by reweighting the loss at each vertex based on its mean contact frequency across the dataset. Please refer to the main paper for further details on the VCB loss.


\begin{table*}[htbp]
\centering
\caption{\textbf{Comparison with SOTA methods of dense hand contact estimation on HIC~\cite{tzionas2016capturing}, RICH~\cite{huang2022capturing}, Hi4D~\cite{yin2023hi4d} dataset.}}
\label{tab:supp_sota_orig_more}
\scalebox{0.87}{
\begin{tabular}{llccc}
\toprule
Test dataset & Methods & $\text{Precision}{\uparrow}$ & $\text{Recall}{\uparrow}$ & $\text{F1-Score}{\uparrow}$ \\
\midrule
\multirow{4}{*}{HIC~\cite{tzionas2016capturing}} & POSA~\cite{hassan2021populating} & 0.000 & 0.000 & 0.000 \\
& BSTRO~\cite{huang2022capturing} & 0.000 & 0.000 & 0.000 \\
& DECO~\cite{tripathi2023deco} & \underline{0.005} & \underline{0.037} & \underline{0.006} \\
& HACO~(Ours) & \textbf{0.216} & \textbf{0.409} & \textbf{0.263} \\
\midrule
\multirow{4}{*}{RICH~\cite{huang2022capturing}} & POSA~\cite{hassan2021populating} & 0.143 & 0.175 & 0.125 \\
& BSTRO~\cite{huang2022capturing} & \underline{0.437} & \underline{0.498} & \underline{0.455} \\
& DECO~\cite{tripathi2023deco} & 0.324 & 0.351 & 0.303 \\
& HACO~(Ours) & \textbf{0.741} & \textbf{0.899} & \textbf{0.781} \\
\midrule
\multirow{4}{*}{Hi4D~\cite{yin2023hi4d}} & POSA~\cite{hassan2021populating} & 0.128 & 0.109 & 0.111 \\
& BSTRO~\cite{huang2022capturing} & \underline{0.313} & \underline{0.207} & \underline{0.247} \\
& DECO~\cite{tripathi2023deco} & 0.197 & 0.162 & 0.146 \\
& HACO~(Ours) & \textbf{0.555} & \textbf{0.636} & \textbf{0.565} \\
\bottomrule
\end{tabular}}
\end{table*}

\subsection{More comparisons with state-of-the-art methods}
\label{sec:more_sota_comparison}
\noindent\textbf{Dense hand contact estimation models.}
In our main manuscript and Table~\ref{tab:supp_sota_orig_more}, we provided comparison between our HACO and state-of-the-art~(SOTA) methods~\cite{hassan2021populating, huang2022capturing, tripathi2023deco} without any modification on model and its model weights.
Nevertheless, the scale of training dataset is largely different between HACO and the SOTA methods.
Also, all of the prior methods are dense human-scene contact estimation methods based on human body models such as SMPL~\cite{loper2015smpl} and SMPL-X~\cite{pavlakos2019expressive}.
To provide the comparison that excludes the effect of large-scale training and discrepancy of performance rooted by different human models~(e.g., SMPL, SMPL-X, MANO), we further provide comparison with SOTA methods that are trained on the same 14 datasets like HACO and modified as hand version in Table~\ref{tab:supp_sota_hand_ver}.
To modify the methods to hand version, we replaced all components of SMPL or SMPL-X modules within POSA~\cite{hassan2021populating}, BSTRO~\cite{huang2022capturing}, DECO~\cite{tripathi2023deco} into MANO~\cite{romero2017embodied} modules and changed corresponding model configurations such as the number of vertices accordingly.
For POSA, as the method requires estimated 3D mesh, we provide the estimated 3D hand mesh with HaMeR~\cite{pavlakos2024reconstructing}.
According to Table~\ref{tab:supp_sota_hand_ver}, HACO still consistently outperforms all hand-specific variants of the full-body methods. 
These results highlight the strength of HACO’s task-specific design based on balanced contact sampling and vertex-level class-balanced loss.

\begin{table*}[htbp]
\centering
\caption{\textbf{Comparison with SOTA methods of dense hand contact estimation with hand version of POSA~\cite{hassan2021populating}, BSTRO~\cite{huang2022capturing}, DECO~\cite{tripathi2023deco} on MOW~\cite{cao2021reconstructing}, HIC~\cite{tzionas2016capturing}, RICH~\cite{huang2022capturing}, Hi4D~\cite{yin2023hi4d} dataset.}}
\label{tab:supp_sota_hand_ver}
\scalebox{0.87}{
\begin{tabular}{llccc}
\toprule
Test dataset & Methods & $\text{Precision}{\uparrow}$ & $\text{Recall}{\uparrow}$ & $\text{F1-Score}{\uparrow}$ \\
\midrule
\multirow{4}{*}{MOW~\cite{cao2021reconstructing}} & POSA~\cite{hassan2021populating} (hand ver. w/ HaMeR~\cite{pavlakos2024reconstructing} pred. mesh) & 0.504 & \underline{0.400} & \underline{0.408} \\
& BSTRO~\cite{huang2022capturing} (hand ver.) & \underline{0.518} & 0.359 & 0.380 \\
& DECO~\cite{tripathi2023deco} (hand ver.) & 0.495 & 0.385 & 0.395 \\
& HACO~(Ours) & \textbf{0.525} & \textbf{0.607} & \textbf{0.522} \\
\midrule
\multirow{4}{*}{HIC~\cite{tzionas2016capturing}} & POSA~\cite{hassan2021populating} (hand ver. w/ HaMeR~\cite{pavlakos2024reconstructing} pred. mesh) & 0.060 & 0.225 & 0.086 \\
& BSTRO~\cite{huang2022capturing} (hand ver.) & \underline{0.070} & \underline{0.293} & \underline{0.107} \\
& DECO~\cite{tripathi2023deco} (hand ver.) & 0.067 & 0.210 & 0.094 \\
& HACO~(Ours) & \textbf{0.216} & \textbf{0.409} & \textbf{0.263} \\
\midrule
\multirow{4}{*}{RICH~\cite{huang2022capturing}} & POSA~\cite{hassan2021populating} (hand ver. w/ HaMeR~\cite{pavlakos2024reconstructing} pred. mesh) & 0.538 & 0.355 & 0.398 \\
& BSTRO~\cite{huang2022capturing} (hand ver.) & \underline{0.581} & \underline{0.413} & \underline{0.421} \\
& DECO~\cite{tripathi2023deco} (hand ver.) & 0.519 & 0.321 & 0.367 \\
& HACO~(Ours) & \textbf{0.741} & \textbf{0.899} & \textbf{0.781} \\
\midrule
\multirow{4}{*}{Hi4D~\cite{yin2023hi4d}} & POSA~\cite{hassan2021populating} (hand ver. w/ HaMeR~\cite{pavlakos2024reconstructing} pred. mesh) & 0.537 & 0.257 & 0.315 \\
& BSTRO~\cite{huang2022capturing} (hand ver.) & \textbf{0.594} & \underline{0.375} & \underline{0.420} \\
& DECO~\cite{tripathi2023deco} (hand ver.) & 0.553 & 0.244 & 0.304 \\
& HACO~(Ours) & \underline{0.555} & \textbf{0.636} & \textbf{0.565} \\
\bottomrule
\end{tabular}}
\end{table*}

\noindent\textbf{Dense hand contact estimation modules.}
While not designed as a standalone dense hand contact estimation model, there are several works that present dense hand contact estimation module as part of 3D reconstruction model~\cite{shimada2023decaf, hu2024learning, wu2024dice, liu2024easyhoi} or pseudo ground-truth generation of dense hand contact~\cite{prakash2025synthesizing}.
For contact modules in 3D reconstruction model, they can be divided into learning-based modules~\cite{shimada2023decaf, hu2024learning, wu2024dice} and heuristic-driven techniques~\cite{liu2024easyhoi, prakash2025synthesizing}.
These methods were excluded from our SOTA comparison in the main manuscript due to the lack of official code release~\cite{shimada2023decaf, hu2024learning, wu2024dice}, the task-specific scope of their training data~\cite{shimada2023decaf, wu2024dice}, and the heuristic nature of their modules~\cite{liu2024easyhoi, prakash2025synthesizing}.
However, we include a broader evaluation in Table~\ref{tab:supp_sota_hand_more} to provide a more comprehensive perspective for future research and to facilitate extensive assessment of hand-specific methods.
We first reproduce DefConNet from Decaf~\cite{shimada2023decaf}, Stage 1 of CHOI~\cite{hu2024learning}, and InteractionNet from DICE~\cite{wu2024dice} based on the model specifications described in their manuscripts, as neither the code nor the pretrained weights are publicly available.
For the Data Engine in LatentAct~\cite{prakash2025synthesizing} and EasyHOI~\cite{liu2024easyhoi}, we use the officially released code to generate outputs for evaluation.
For all learning-based modules~\cite{shimada2023decaf, hu2024learning, wu2024dice}, we train the reproduced modules based on 14 datasets like HACO but follow their training configurations such as loss function or optimizers when provided in their manuscripts.
In Table~\ref{tab:supp_sota_hand_more}, the results show that our HACO outperforms all previous dense hand contact estimation modules by a significant margin.
This highlights the need for a standalone dense hand contact estimation model specifically trained for hand contact, and demonstrates the effectiveness of HACO’s task-specific design.

\begin{table*}[htbp]
\centering
\caption{\textbf{Comparison with SOTA methods of dense hand contact estimation with hand contact estimation modules (†: reproduced, re-trained with 14 datasets, code not available) on MOW~\cite{cao2021reconstructing} dataset.}}
\label{tab:supp_sota_hand_more}
\scalebox{0.87}{
\begin{tabular}{lccc}
\toprule
Methods & $\text{Precision}{\uparrow}$ & $\text{Recall}{\uparrow}$ & $\text{F1-Score}{\uparrow}$ \\
\midrule
†DefConNet (from Decaf~\cite{shimada2023decaf}) & 0.414 & \underline{0.404} & 0.368 \\
†Stage 1 in CHOI~\cite{hu2024learning} & \underline{0.521} & 0.387 & 0.407 \\
†InteractionNet (from DICE~\cite{wu2024dice}) & 0.459 & 0.392 & \underline{0.413} \\
Data Engine in LatentAct~\cite{prakash2025synthesizing} (w/ SAM2~\cite{ravi2024sam} \& HaMeR~\cite{pavlakos2024reconstructing}) & 0.345 & 0.244 & 0.211 \\
EasyHOI~\cite{liu2024easyhoi} & 0.480 & 0.228 & 0.282 \\
\midrule
HACO~(Ours) & \textbf{0.525} & \textbf{0.607} & \textbf{0.522} \\
\bottomrule
\end{tabular}}
\end{table*}

\subsection{More ablation studies}
\label{sec:more_abl_study}
\noindent\textbf{Effectiveness of training dataset size.}
To assess the contribution of large-scale training to HACO’s performance, we conducted an ablation study comparing different training dataset sizes spanning 1, 3, and 14 datasets.
The results demonstrate that performance improves notably as the training dataset size increases, emphasizing the importance of large-scale and diverse data for robust contact estimation.
In particular, training on the full set of 14 datasets yields the best performance, while using only three representative datasets of DexYCB~\cite{chao2021dexycb}, ObMan~\cite{hasson2019learning}, MOW~\cite{cao2021reconstructing} and even one dataset of MOW dataset still produces competitive results, indicating that HACO remains effective even with limited training dataset size.

\begin{table*}[htbp]
\centering
\caption{\textbf{Ablation for training dataset size on MOW~\cite{cao2021reconstructing} dataset.}}
\label{tab:supp_abl_train_size}
\scalebox{0.87}{
\begin{tabular}{cccc}
\toprule
Methods & $\text{Precision}{\uparrow}$ & $\text{Recall}{\uparrow}$ & $\text{F1-Score}{\uparrow}$ \\
\midrule
HACO trained on 1 dataset & \underline{0.498} & 0.348 & 0.373 \\
HACO trained on 3 datasets & 0.463 & \underline{0.602} & \underline{0.485} \\
HACO trained on 14 datasets & \textbf{0.525} & \textbf{0.607} & \textbf{0.522} \\
\bottomrule
\end{tabular}}
\end{table*}

\subsection{Quantitative results with different backbones}
\label{sec:more_backbones}
We evaluate the performance of HACO with various backbone architectures, keeping the rest of the pipeline fixed. 
As shown in Table~\ref{tab:supp_diff_back}, our original model using ViT-H~\cite{dosovitskiy2020image} pretrained on HaMeR~\cite{pavlakos2024reconstructing} achieves the highest F1-score, demonstrating the strongest overall performance. 
ViT-based backbones consistently outperform convolutional alternatives, with ViT-B achieving the second highest recall and ViT-S maintaining strong precision. 
ViT-L also performs competitively, achieving a strong F1-score.
This highlights the effectiveness of Transformer~\cite{vaswani2017attention}-based architectures for dense contact estimation, likely due to their global receptive fields and capacity to model long-range dependencies.
To assess the benefit of pretraining on related tasks, we additionally train and evaluate a variant with FPN~\cite{lin2017feature} backbone pretrained with HandOccNet~\cite{park2022handoccnet}, a 3D hand mesh reconstruction model. 
This configuration yields low F1-score compared to ImageNet-pretrained backbones, suggesting that reconstruction-specific pretraining alone is insufficient for accurate hand contact prediction. 
HRNet backbones, widely used in 3D human mesh reconstruction~\cite{lin2021end, lin2021mesh, li2022cliff, zhang2023pymaf, li2025hybrik}, achieve moderate performance in our setting, with F1-scores below 0.5 despite their high-resolution design. 
ResNet backbones~\cite{he2016deep} exhibit a wide performance range, with deeper variants such as ResNet-152 performing relatively well, while lighter variants such as ResNet-18 offer improved efficiency at the cost of reduced F1-score.
Overall, these results indicate that the performance gains of our final HACO model arise from the combination of the strong representational capacity of the ViT backbone and task-specific knowledge transferred from 3D hand mesh reconstruction.

\begin{table*}[htbp]
\centering
\caption{\textbf{Comparison of various backbone models on MOW~\cite{cao2021reconstructing} dataset.}}
\scalebox{0.87}{\begin{tabular}{lcccccc} \toprule
Model & Backbone & Pretrained & $\text{Precision}{\uparrow}$ & $\text{Recall}{\uparrow}$ & $\text{F1-Score}{\uparrow}$ \\
\midrule
HACO & ViT-H~\cite{dosovitskiy2020image} & HaMeR~\cite{pavlakos2024reconstructing} & \textbf{0.525} & \textbf{0.607} & \textbf{0.522} \\
HACO & FPN~\cite{lin2017feature} & HandOccNet~\cite{park2022handoccnet} & 0.505 & 0.575 & 0.482 \\
\midrule
HACO & ViT-L~\cite{dosovitskiy2020image} & ImageNet~\cite{deng2009imagenet} & 0.493 & 0.598 & \underline{0.491} \\
HACO & ViT-B~\cite{dosovitskiy2020image} & ImageNet~\cite{deng2009imagenet} & 0.488 & \underline{0.604} & 0.484 \\
HACO & ViT-S~\cite{dosovitskiy2020image} & ImageNet~\cite{deng2009imagenet} & \underline{0.510} & 0.514 & 0.462 \\
\midrule
HACO & HRNet-W48~\cite{sun2019deep} & ImageNet~\cite{deng2009imagenet} & 0.509 & 0.563 & 0.485 \\
HACO & HRNet-W32~\cite{sun2019deep} & ImageNet~\cite{deng2009imagenet} & 0.500 & 0.558 & 0.471 \\
\midrule
HACO & ResNet-152~\cite{he2016deep} & ImageNet~\cite{deng2009imagenet} & 0.506 & 0.558 & 0.486 \\
HACO & ResNet-101~\cite{he2016deep} & ImageNet~\cite{deng2009imagenet} & 0.498 & 0.564 & 0.480 \\
HACO & ResNet-50~\cite{he2016deep} & ImageNet~\cite{deng2009imagenet} & 0.490 & 0.558 & 0.470 \\
HACO & ResNet-34~\cite{he2016deep} & ImageNet~\cite{deng2009imagenet} & 0.489 & 0.558 & 0.466 \\
HACO & ResNet-18~\cite{he2016deep} & ImageNet~\cite{deng2009imagenet} & 0.484 & 0.538 & 0.456 \\

\bottomrule
\end{tabular}}
\vspace{-0.3cm}
\label{tab:supp_diff_back}
\end{table*}

\subsection{Computational requirements}
\label{sec:computation}
Table~\ref{tab:supp_comp_require} reports the computational requirements of HACO with various backbone configurations. 
The original HACO model, listed in the first row, shows the highest computational cost, particularly in training memory, which exceeds 26GB. 
This is primarily due to the memory overhead associated with storing gradients while fine-tuning the ViT-H~\cite{dosovitskiy2020image} backbone pretrained on HaMeR~\cite{pavlakos2024reconstructing}.
To support broader adoption in downstream tasks such as 3D hand grasp generation~\cite{grady2021contactopt} and 3D hand-object reconstruction~\cite{liu2024easyhoi}, we provide additional HACO variants with reduced computational demand. 
We first evaluate the FPN~\cite{lin2017feature} backbone from HandOccNet~\cite{park2022handoccnet}. 
Although FPN is known to be computationally expensive due to its multi-scale feature hierarchy, it remains significantly more efficient than ViT-H. 
We also test lighter ViT variants such as ViT-B and ViT-S, which offer favorable trade-offs by maintaining moderate memory usage while achieving inference speeds exceeding 60 fps.
We further evaluate HRNet~\cite{sun2019deep} backbones, including HRNet-W48 and HRNet-W32. 
Despite their moderate memory consumption, both models exhibit low inference speeds, falling below 20 fps. 
Finally, we include ResNet-based~\cite{he2016deep} variants ranging from lightweight models such as ResNet-18 to deeper configurations such as ResNet-152. 
These models generally show low memory usage, fewer parameters, and high runtime efficiency. 
In particular, ResNet-18 achieves an inference speed exceeding 100 fps while maintaining a computational cost under 3 GFLOPs.

\begin{table*}[htbp]
\centering
\setlength{\tabcolsep}{4pt}  
\caption{\textbf{Computational requirements of various backbone configurations.}}
\scalebox{0.87}{\begin{tabular}{lcccccc} \toprule
Model & Backbone & Train Memory~(MB) & Test Memory~(MB) & Params.~(M) & Speed~(fps) & GFLOPs \\
\midrule
HACO & ViT-H~\cite{xu2022vitpose} & 26,157 & 12,728 & 671.10 & 54.26 & 125.63 \\
HACO & FPN~\cite{lin2017feature} & 6,887 & 3,002 & 59.56 & 66.71 & 9.94 \\
\midrule
HACO & ViT-L~\cite{xu2022vitpose} & 16,649 & 4,604 & 341.91 & 65.79 & 62.87\\
HACO & ViT-B~\cite{xu2022vitpose} & 7,767 & 2,940 & 122.84 & 96.80 & 18.54 \\
HACO & ViT-S~\cite{xu2022vitpose} & 4,960 & 2,434 & 56.34 & 86.88 & 5.10 \\
\midrule
HACO & HRNet-W48~\cite{sun2019deep} & 11,296 & 7,574 & 122.37 & 10.52 & 23.48 \\
HACO & HRNet-W32~\cite{sun2019deep} & 9,026 & 7,806 & 86.13 & 17.14 & 12.55 \\
\midrule
HACO & ResNet-152~\cite{he2016deep} & 9,252 & 5,096 & 103.05 & 29.84 & 15.93 \\
HACO & ResNet-101~\cite{he2016deep} & 7,471 & 4,978 & 87.40 & 41.14 & 11.07 \\
HACO & ResNet-50~\cite{he2016deep} & 5,919 & 4,826 & 68.41 & 73.25 & 6.21 \\
HACO & ResNet-34~\cite{he2016deep} & 4,163 & 2,456 & 56.75 & 80.85 & 5.03 \\
HACO & ResNet-18~\cite{he2016deep} & 3,785 & 2,372 & 46.64 & 111.23 & 2.61 \\

\bottomrule
\end{tabular}}
\vspace{-0.3cm}
\label{tab:supp_comp_require}
\end{table*}

\subsection{More qualitative results}
\label{sec:more_qualitative}
\begin{figure*}[htbp]
\begin{center}
\includegraphics[width=1.0\linewidth]{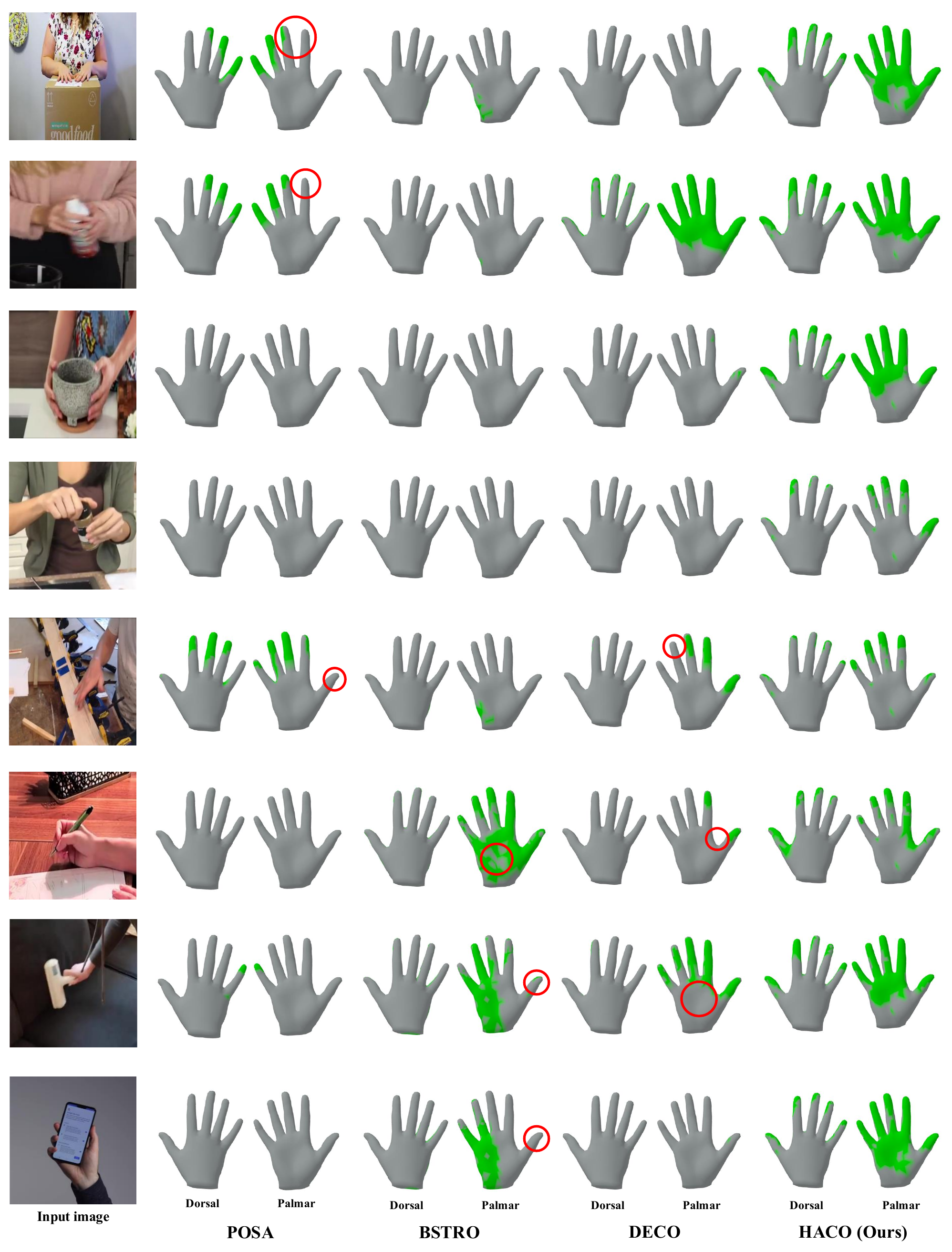}
\end{center}
\caption{
\textbf{Qualitative comparison of dense hand contact estimation with POSA~\cite{hassan2021populating}, BSTRO~\cite{huang2022capturing}, DECO~\cite{tripathi2023deco} on MOW~\cite{cao2021reconstructing} dataset.} We highlight exemplar regions where HACO outperforms previous methods.
Note that we only predict right hand contact.}
\label{fig:supp_hand_sota_contact_qual_mow}
\end{figure*}

\begin{figure*}[htbp]
\begin{center}
\includegraphics[width=1.0\linewidth]{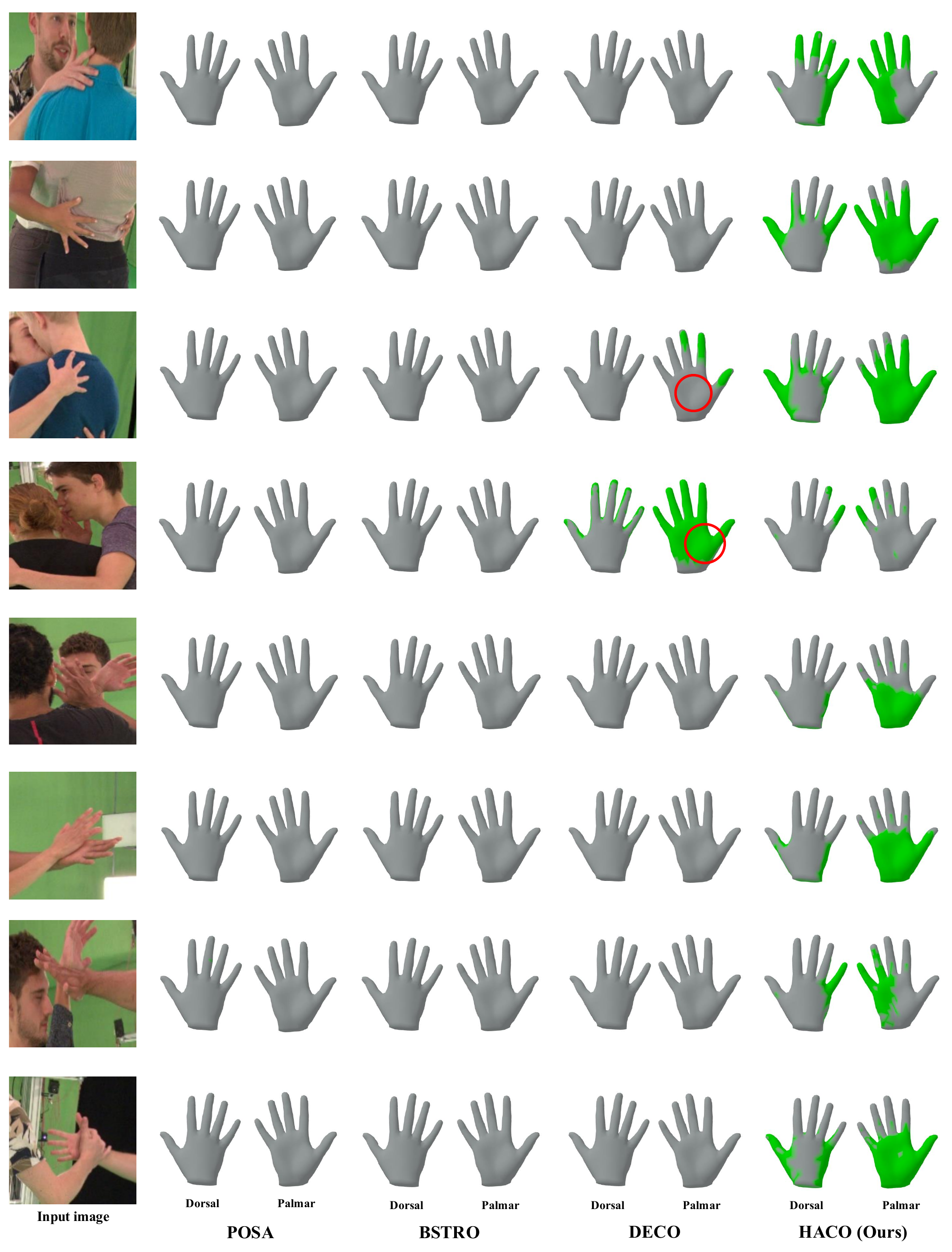}
\end{center}
\caption{
\textbf{Qualitative comparison of dense hand contact estimation with POSA~\cite{hassan2021populating}, BSTRO~\cite{huang2022capturing}, DECO~\cite{tripathi2023deco} on Hi4D~\cite{yin2023hi4d} dataset.} We highlight exemplar regions where HACO outperforms previous methods.
Note that we only predict right hand contact.}
\label{fig:supp_hand_sota_contact_qual_hi4d}
\end{figure*}

\begin{figure*}[htbp]
\begin{center}
\includegraphics[width=1.0\linewidth]{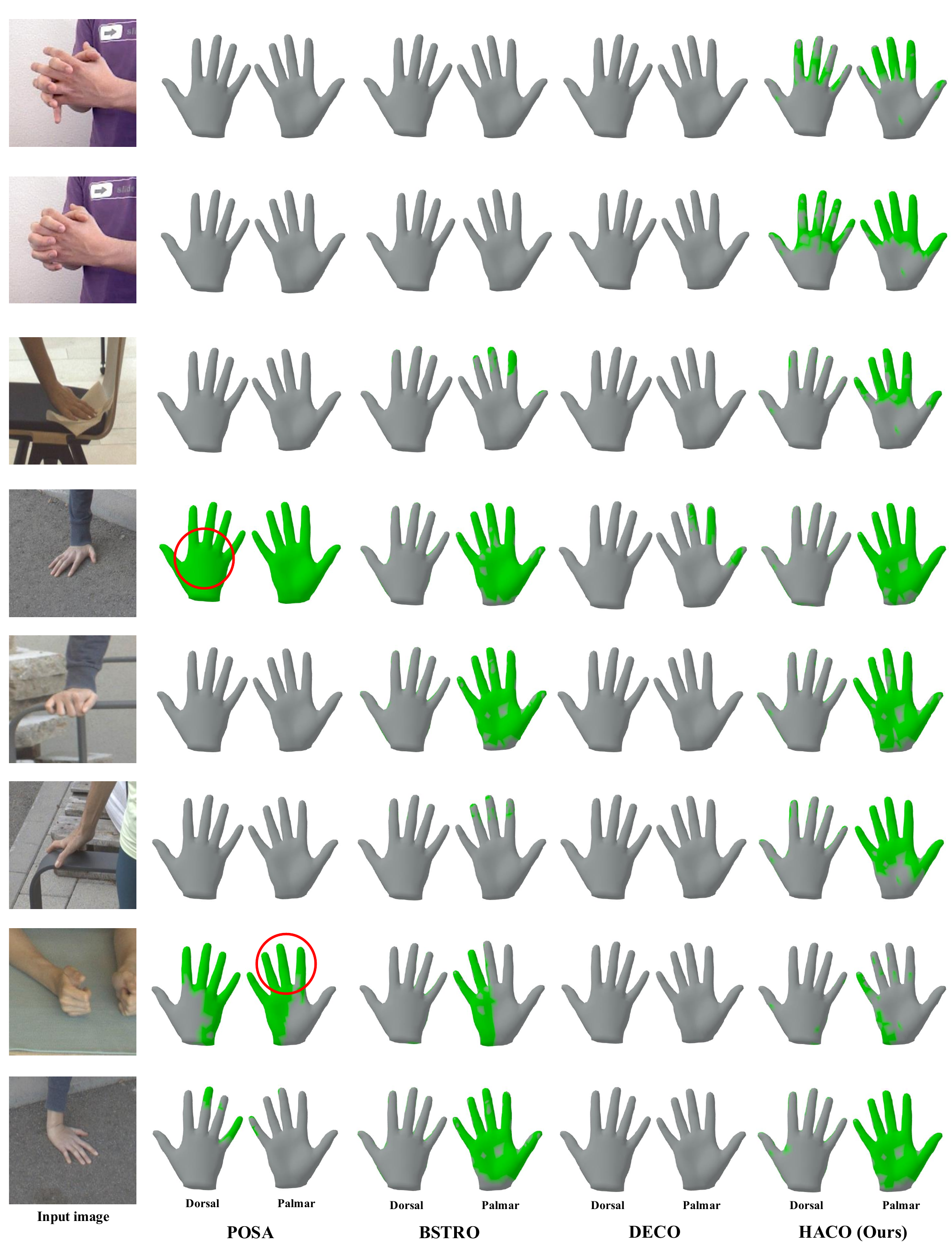}
\end{center}
\caption{
\textbf{Qualitative comparison of dense hand contact estimation with POSA~\cite{hassan2021populating}, BSTRO~\cite{huang2022capturing}, DECO~\cite{tripathi2023deco} on HIC~\cite{tzionas2016capturing} and RICH~\cite{huang2022capturing} dataset.} We highlight exemplar regions where HACO outperforms previous methods.
Note that we only predict right hand contact.}
\label{fig:supp_hand_sota_contact_qual_hic_rich}
\end{figure*}

Additional qualitative comparisons of dense hand contact estimation between HACO and prior methods, including POSA~\cite{hassan2021populating}, BSTRO~\cite{huang2022capturing}, and DECO~\cite{tripathi2023deco}, are shown in Figures~\ref{fig:supp_hand_sota_contact_qual_mow}, \ref{fig:supp_hand_sota_contact_qual_hi4d}, and \ref{fig:supp_hand_sota_contact_qual_hic_rich}, which correspond to the MOW~\cite{cao2021reconstructing}, Hi4D~\cite{yin2023hi4d}, and both HIC~\cite{tzionas2016capturing} and RICH~\cite{huang2022capturing} datasets, respectively.
Overall, HACO consistently outperforms existing approaches by a substantial margin. 
Unlike prior methods that frequently fail to predict contact even when the hand is clearly interacting, HACO reliably captures accurate dense hand contact.
Among the baselines, DECO demonstrates stronger performance than POSA and BSTRO, likely as a result of being trained on in-the-wild datasets with more diverse contact distributions.
However, it remains to be affected by class imbalance, resulting in a high rate of false negatives. 
BSTRO is more severely affected by this imbalance, frequently predicting no contact across the entire hand and producing overly coarse contact regions concentrated in the palmar area. 
This behavior likely stems from its reliance on human-scene interaction datasets during training. 
POSA also exhibits frequent false negatives and fails to detect contact even in visibly interacting regions, likely due to its dependence on a pose prior rather than direct contact supervision with visual input.


\subsection{Limitations and societal impacts}
\label{sec:limitations}

\noindent\textbf{Limitations.}
Our HACO supports dense hand contact estimation across a wide range of interaction types, including hand-object, hand-hand, hand-face, hand-scene, and hand-body interactions. 
However, hands frequently engage in self-contact under certain poses (e.g., contact between the thumb and index finger in an "okay" gesture) or during whole-body motions~\cite{muller2021self}. 
We intentionally exclude self-contact cases, as including them could negatively impact downstream applications. 
For example, ContactOpt~\cite{grady2021contactopt} may incorrectly optimize grasps between the hand and an object if self-contact is misinterpreted as external contact. 
As a result, HACO currently does not support dense hand contact estimation in self-contact scenarios. 
Nevertheless, we consider modeling self-contact a promising direction for future research, particularly for applications in the metaverse (e.g., AR/VR) and action recognition.
Furthermore, temporal context is essential to learn accurate dense hand contact estimation as contact is most noticeable when the movement of hand is obstructed by an object or scene.
Integration of temporal-based methods~\cite{zhao2023poseformerv2, nam2023cyclic} that facilitate temporal information from input video can stabilize the performance of dense hand contact estimation.
With the advent of diffusion-based methods~\cite{lee2025semanticdraw, lu2025dposer}, designing a diffusion-based approach for dense hand contact estimation is another important research direction as it enables the learning of dense hand contact priors from the large-scale data.
Additionally, our balanced contact sampling (BCS) utilizes a single reference point of dataset-wide dense hand contact mean.
While this allows HACO to fairly represent diverse contact statistics, if distribution of the statistics is multi-modal, this strategy might not be effective.
There can be a room for improvement on our BCS strategy if the multi-modality of dense hand contact datasets is proven and multi-modal aware methodology can be incorporated to BCS.
Moreover, we evaluate HACO only on samples that contain at least one vertex in contact.
However, this evaluation protocol may overlook the model’s inability to handle false hand-level contact predictions, which are important to consider for real-world applications.
The incorporation of additional evaluation protocols such as accuracy, specificity, and Matthews Correlation Coefficient may be beneficial.
Lastly, dense hand contact estimation demonstrates inferior performance on interacting hand scenarios relative to other hand interaction scenarios. 
The incorporation of 3D interacting hand mesh reconstruction method~\cite{fan2021learning, park2023extract} or the use of additional training dataset for hand-hand interaction~\cite{moon2023dataset} could enhance the performance.

\noindent\textbf{Societal impacts.}
The proposed method has broad potential for applications involving hand interaction analysis, including AR/VR, robotics, and behavioral understanding. 
However, given the inherently human-centered nature of the task, there is a risk of misuse in areas such as mass surveillance or privacy infringement. 
We strongly discourage the deployment of this system in applications that may compromise human rights or personal privacy, and urge practitioners to consider ethical implications when applying HACO to downstream tasks.

\end{document}